\documentclass[lettersize,journal]{IEEEtran}
\usepackage{amsmath,amsfonts}
\usepackage{algorithmic}
\usepackage{algorithm}
\usepackage{array}
\usepackage[caption=false,font=normalsize,labelfont=sf,textfont=sf]{subfig}
\usepackage{textcomp}
\usepackage{stfloats}
\usepackage{url}
\usepackage{verbatim}
\usepackage{graphicx}
\usepackage{cite}
\usepackage[version=4]{mhchem}
\usepackage{graphicx} 
\usepackage{epsfig} 
\usepackage{mathptmx} 
\usepackage{times} 
\usepackage{amsmath} 
\usepackage{amssymb} 
\usepackage{mathrsfs}
\usepackage{bbding}

\usepackage[T1]{fontenc}
\usepackage{tabularx}  
\usepackage{hyperref}
\hypersetup{hidelinks,
	colorlinks=true,
	allcolors=black,
	pdfstartview=Fit,
	breaklinks=true}
\usepackage{multirow}
\usepackage{url}
\usepackage{makecell}
\usepackage{diagbox} 
\begin{document}
\title{Visual-tactile Fusion for Transparent Object Grasping in Complex Backgrounds}
\author{Shoujie Li,~\IEEEmembership{Student Member,~IEEE}, Haixin Yu, Wenbo Ding,~\IEEEmembership{Member,~IEEE}, Houde Liu,~\IEEEmembership{Member,~IEEE}, Linqi Ye,~\IEEEmembership{Member,~IEEE}, Chongkun Xia,~\IEEEmembership{Member,~IEEE}, Xueqian Wang,~\IEEEmembership{Member,~IEEE}, Xiao-Ping Zhang,~\IEEEmembership{Fellow,~IEEE}
	\thanks{This work was supported by the National Key R\&D Program of China (2022YFB4701400/4701402),   Shenzhen Science and Technology Program (JCYJ20220530143013030), Guangdong Innovative and Entrepreneurial Research Team Program (2021ZT09L197),  Shenzhen Science Fund for Distinguished Young Scholars (RCJC20210706091946001),  Guangdong Special Branch Plan for Young Talent with Scientiﬁc and Technological Innovation (2019TQ05Z111), National Natural Science Foundation of China (92248304, U21B6002, 62203260, 62003188, 62104125), Natural Science Foundation of Guangdong Province (2023A1515011773), China Postdoctoral Science Foundation (2022M711823), Tsinghua Shenzhen International Graduate School-Shenzhen Pengrui Young Faculty Program of Shenzhen Pengrui Foundation (No. SZPR2023005). \textit{(Shoujie Li and Haixin Yu contributed equally to this
work.) (Corresponding author: Wenbo Ding \& Houde Liu, ding.wenbo@sz.tsinghua.edu.cn  \&  liu.hd@sz.tsinghua.edu.cn)}}   %
	\thanks{Shoujie Li, Haixin Yu, Wenbo Ding, Houde Liu, Chongkun Xia, Xueqian Wang, Xiao-Ping Zhang are with Tsinghua Shenzhen International Graduate School,  Shenzhen 518055, China.}
	\thanks{Shoujie Li, Wenbo Ding are also with the RISC-V International Open Source Laboratory, Tsinghua-Berkeley Shenzhen Institute, Shenzhen 518055, China.}
	\thanks{Linqi Ye is with the  Institute of Artificial Intelligence, Collaborative Innovation Center for the Marine Artificial Intelligence, Shanghai University, Shanghai 200444, China.}
	\thanks{Xiao-Ping Zhang is also with the Department of Electrical, Computer, and Biomedical Engineering, Ryerson University, Toronto, ON M5B 2K3, Canada.}
	\thanks{This paper has supplementary downloadable material available at \href{https://sites.google.com/view/visual-tactilefusion}{https://sites.google.com/view/visual-tactilefusion}.}
}

\markboth{IEEE Transactions on Robotics}%
{IEEE Transactions on Robotics}

\maketitle
\begin{abstract}
The grasping of transparent objects is challenging but of significance to robots. In this article, a visual-tactile fusion framework for transparent object grasping in complex backgrounds is proposed, which synergizes the advantages of vision and touch, and greatly improves the grasping efficiency of transparent objects. First, we propose a multi-scene synthetic grasping dataset named SimTrans12K together with a Gaussian-Mask annotation method. Next, based on the TaTa gripper, we propose a grasping network named transparent object grasping convolutional neural network (TGCNN) for grasping position detection, which shows good performance in both synthetic and real scenes. Inspired by human grasping, a tactile calibration method and a visual-tactile fusion classification method are designed, which improve the grasping success rate by 36.7\% compared to direct grasping and the classification accuracy by 39.1\%. Furthermore, the Tactile Height Sensing (THS) module and the Tactile Position Exploration (TPE) module are added to solve the problem of grasping transparent objects in irregular and visually undetectable scenes. Experimental results demonstrate the validity of the framework.
\end{abstract}


\section{Introduction}
Transparent objects are common in people's daily life, but it is very challenging for robots to accurately detect and grasp them. This is mainly because the appearance of transparent objects changes drastically under different backgrounds, making traditional visual detection prone to failure. Therefore, how to realize accurate and robust detection of transparent objects towards efficient grasp has attracted tremendous interest in the field of robotics. Representative works include the multi-modal transfer learning method~\cite{weng2020multi}, the plenoptic sensing approach~\cite{zhou2019glassloc}, the transparent depth information complimenting method~\cite{sajjan2020clear}, etc. Nevertheless, these methods usually focus on the detection of transparent objects and assume the objects are placed in static backgrounds with simple patterns, which is not always the case in practice. Hence, it is of great significance to develop a grasping method for transparent objects that can adapt to various backgrounds, e.g., the objects placed on soft or fluid surfaces, with complex patterns or unpredictable conditions, such as undulating scenes, underwater, and so on.

\begin{figure}
	\centering
	\includegraphics[width=0.48\textwidth]{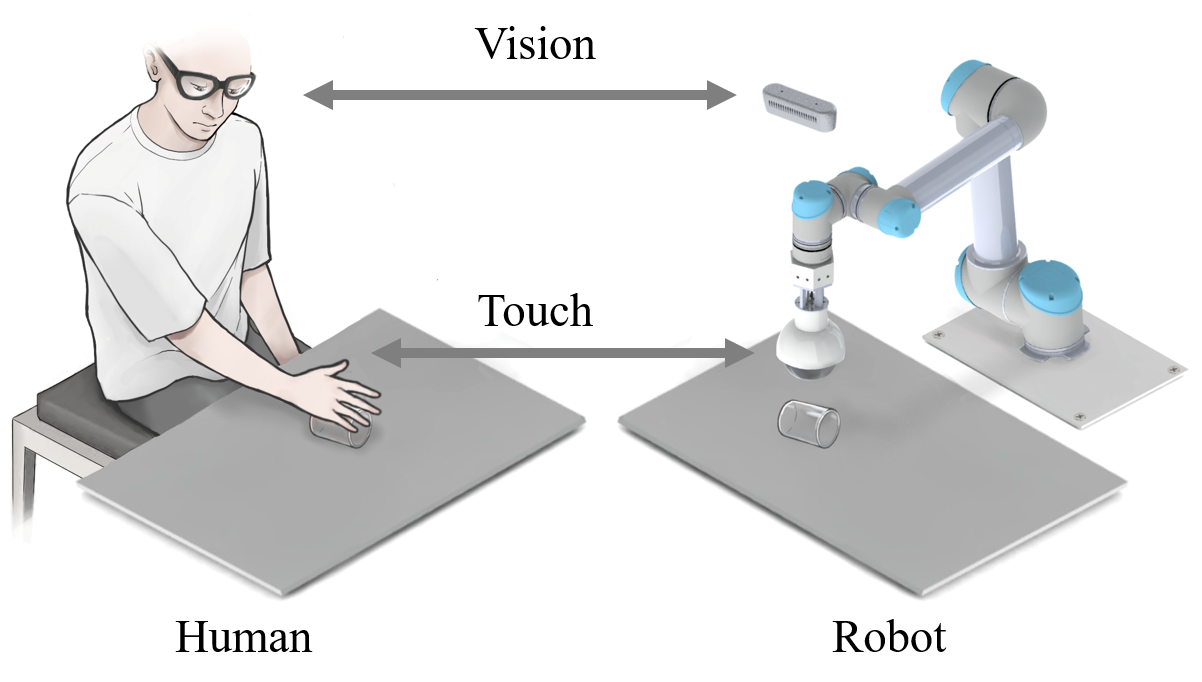}
	\caption{The visual-tactile fusion framework inspired by human grasping.} \label{fig:1}
\end{figure}
Thanks to the development of computer vision and deep learning, vision-assisted perception has now become a popular and effective choice for robot interactions and environment explorations. However, the vision-based method cannot work well in dim, reflective, and cloudy conditions. Inspired by the grasping behavior of humans shown in Fig.~\ref{fig:1}, where visual and tactile sensation are collaboratively working towards complicated tasks, a visual-tactile fusion-based framework using the TaTa gripper~\cite{9811806} is proposed in this paper for transparent object grasping in complex backgrounds. Here the tactile sensation is utilized to compensate for the limitation of vision, which not only largely raises the success rate of grasping by 36.7\% but also greatly improves the classification accuracy of transparent objects by 39.1\%. In addition, the framework can be extended to cover more challenging scenes such as irregular backgrounds or even visually undetectable scenes. 
Specifically, the contributions of this work are fourfold:

\begin{itemize}
\item Firstly, a  synthetic transparent object dataset named SimTrans12K is proposed containing different styles of backgrounds, lighting, and camera positions, which has more complex and abundant background information than the previous transparent object datasets, such as ClearGrasp~\cite{sajjan2020clear} and Dex-Nerf~\cite{ichnowski2021dex}. Besides, to improve the performance of Sim2Real, we propose a Gaussian-Mask method for transparent object grasping position annotation, which can better represent the position information of transparent objects than the binary ground truth grasping position~\cite{weng2020multi}.

\item Secondly, for the TaTa gripper~\cite{9811806}, a generative grasping network named transparent object grasping convolutional neural network (TGCNN) is proposed, which can achieve transparent object grasping position detection in complex backgrounds and lighting with training from the synthetic dataset only. Meanwhile, a tactile information extraction algorithm and a visual-tactile fusion-based transparent object classification algorithm are developed to compensate for the visual deviation~\cite{seib2017friend}.

\item Thirdly, to realize transparent object grasping in complex backgrounds, we propose a visual-tactile fusion-based transparent object grasping framework with tactile calibration. Besides, we add the Tactile Height Sensing (THS) module and the Tactile Position Exploration (TPE) module to this framework, which can achieve transparent object grasping in stacking, overlapping, or even visually undetectable scenes. Those scenes are extremely difficult and there are only a few studies before~\cite{dai2022domain, huang2015sensor,chang2021ghostpose,ichnowski2021dex,sajjan2020clear}.

\item Finally, to test the effectiveness of the proposed framework, we carefully design several experiments to extensively compare the performance with several state-of-the-art baseline methods, which indicates the proposed method has a considerable performance improvement for transparent object grasping and classification. Moreover, we also test the proposed method in some highly difficult scenes such as stacking, overlapping, undulating, and dynamic underwater environments, which greatly extends the application areas of transparent object grasping.

\end{itemize}

The rest of this paper is organized as follows. The related work is reviewed in Section II. The hardware setup is detailed in Section III. Section IV presents the synthetic data generation, the grasping position detection algorithm, the tactile information extraction algorithm, and the visual-tactile fusion-based classification algorithm. The proposed visual-tactile fusion grasping strategy is presented in Section V. Furthermore, experimental validations are provided in Section VI. Finally, Section VII concludes this paper.

\section{Related Work}
\subsection{Transparent Object Dataset}

Xie \textit{et al.} proposed a transparent dataset Trans10K with 10,428 real data~\cite{xie2020segmenting}, but it only has two limited categories, which was further refined to 11 fine-grained categories of transparent objects in the dataset Trans10K-v2~\cite{xie2021segmenting}.  Jiang \textit{et al.} constructed a real-world dataset TRANS-AFF with affordances and depth maps of transparent objects~\cite{jiang2022a4t}.
\begin{figure}
	\centering
	\includegraphics[width=0.48\textwidth]{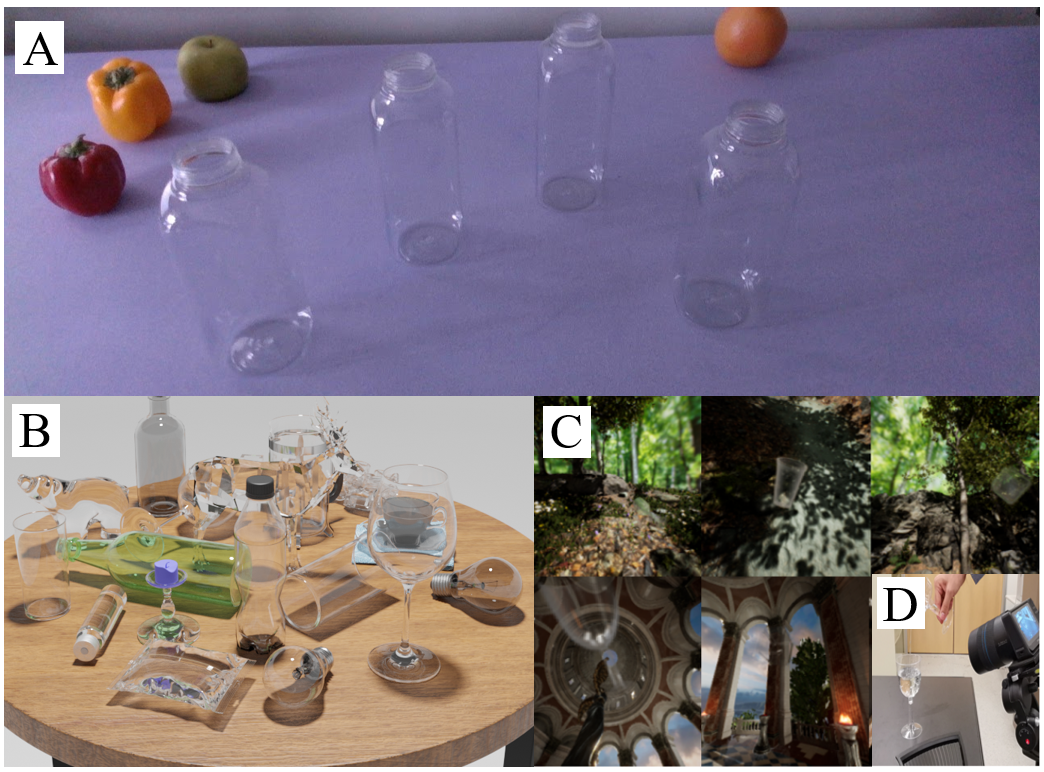}
	\caption{Examples of transparent object dataset. (A) ClearGrasp~\cite{sajjan2020clear}. (B) Dex-NeRF~\cite{ichnowski2021dex}. (C) LIT~\cite{zhou2020lit}. (D) Light Field Camera used in LIT dataset.  } \label{fig:Comparisondata}
\end{figure}

With the development of powerful computer graphics simulation tools, researchers have tried to generate the synthetic dataset of the transparent object from simulation considering its low cost, simplicity, and efficiency. Representative works include ClearGrasp~\cite{sajjan2020clear}, a synthetic dataset for depth-completion tasks, Dex-NeRF~\cite{ichnowski2021dex}, a synthetic dataset for transparent object detection and localization, and LIT~\cite{zhou2020lit}, a synthetic dataset for light-field cameras, as shown in Fig.~\ref{fig:Comparisondata}.

The implementation of transparent object grasping via Sim2Real puts higher demands on the diversity and validity of the dataset, so we hope that the dataset contains more transparent object data under complex backgrounds and lightness, while ClearGrasp and Dex-NeRF can hardly meet such requirements. Although LIT contains more complex scenes, it contains scenes with low brightness  and is designed for light field cameras, as shown in Fig.~\ref{fig:Comparisondata}(D). A new synthetic dataset for the transparent object is proposed in this paper. To reduce the discrepancy between the synthetic data and the real scene, we carefully calibrate the parameters of the cameras, lights, and backgrounds in the simulation software to make them as consistent as possible with the real camera, which is neglected in the existing transparent object synthetic dataset. In addition, we also propose a  Gaussian-Mask annotation method for transparent objects.

\subsection{Transparent Object Detection}
The visual detection methods of transparent objects can be divided into two types: physical feature-based detection methods and deep learning-based methods.

Traditional methods detect transparent objects based on physical features such as deformation, reflection, and image gradient changes. Fritz \textit{et al.} reported an additive latent feature model through the assumption that the texture of transparent objects originates from the background~\cite{fritz2009additive}. McHenry \textit{et al.} proposed a hierarchical support vector machine (SVM)-based glass edge recognition model via the background texture distortion and reflection phenomenon at the glass edge~\cite{mchenry2005finding}. Maeno \textit{et al.} used a light field camera to acquire images and utilized a light field distortion feature (LDF) to describe the distortion caused by the refraction of transparent objects~\cite{maeno2013light}.

The development of deep learning paves a new way for transparent object detection. Liu \textit{et al.} used a convolutional neural network called Single Shot MultiBox Detector (SSD) for transparent object detection~\cite{liu2016ssd}. Xie \textit{et al.} proposed a Transformer-based segmentation pipeline termed Trans2Seg~\cite{xie2020segmenting}. Fan \textit{et al.} applied the transparent object detection to highly dynamic scenes and proposed a recognition tracking network named TransATOM, which can stably track the transparent objects in video~\cite{fan2021transparent}.
Xu \textit{et al.} proposed a joint point cloud and depth completion method, which can complete the depth of transparent objects in cluttered scenes~\cite{xu2021seeing}. Zhu \textit{et al.} presented a novel framework that can complete missing depth given noisy RGB-D inputs~\cite{zhu2021rgb}.

Deep learning methods have demonstrated superior robustness to traditional ones, especially for transparent object detection in complex scenarios, showing great application potential.

\subsection{Transparent Object Grasping}
Transparent object grasping is another challenging task. Apart from the object position, the optimal grasping position and angle should be considered as well during grasping. We classify transparent object grasping tasks into different levels of difficulty as shown in Table~\ref{tabt}, ranging from the simple case of grasping on a plane to the extremely difficult case of grasping in dynamic underwater environments. 

For transparent object grasping, most of the work is performed on  planes with a simple background. For example, Weng \textit{et al.} proposed a multi-modal transfer learning method for transparent and reflective object grasping~\cite{weng2020multi}. Sajjan \textit{et al.} reported a transparent depth completion method to grasp transparent objects~\cite{sajjan2020clear}. Liu \textit{et al.} proposed a keypoint-based method for 6D pose estimation of objects using stereo image input, which can be easily applied to transparent object grasping~\cite{liu2020keypose}. Ichnowski \textit{et al.} rendered depth maps of transparent objects using neural radiation fields (NeRF) to infer the geometry of transparent objects and perform plane grasping~\cite{ichnowski2021dex}. Kerr \textit{et al.} proposed Evolving NeRF (Evo-NeRF), leveraging recent speedups in NeRF training and further extending it to rapidly train the NeRF representation concurrently to image capturing~\cite{kerrevo}. Cao \textit{et al.} proposed a fuzzy-depth soft grasping (FSG) algorithm for Tstone-Soft (TSS) gripper~\cite{cao2021fuzzy}.

Besides, grasping transparent objects in complex scenes, e.g., glass fragments, stacking, overlapping, undulating, sand and underwater scenes are more challenging but of practical meaning.
Firstly, glass fragments are a type of object with no fixed model, and their detection and grasping pose a significant challenge. Because of its random shape and the presence of more angles, the accuracy of grasping and the universality of grasping tools are highly required, so there is almost no research on the grasping of transparent glass fragments. For overlapping and stacking transparent objects, their texture will be merged with the background, so it is difficult to distinguish them. Zhou \textit{et al.} proposed a GlassLoc algorithm for grasping pose detection of transparent objects in clutter using plenoptic sensing, which achieves transparent object grasping in stacking scenes though the experimental setting is simple~\cite{zhou2019glassloc}. Lysenkov \textit{et al.} proposed a method that can achieve pose estimation for cluttered transparent objects in complex backgrounds, but it applies the pose matching method which is only applicable to objects within the dataset~\cite{lysenkov2013pose} and cannot solve the grasping of transparent fragments without regular shapes.

\begin{figure}[htp]
	\centering
	\includegraphics[width=0.48\textwidth]{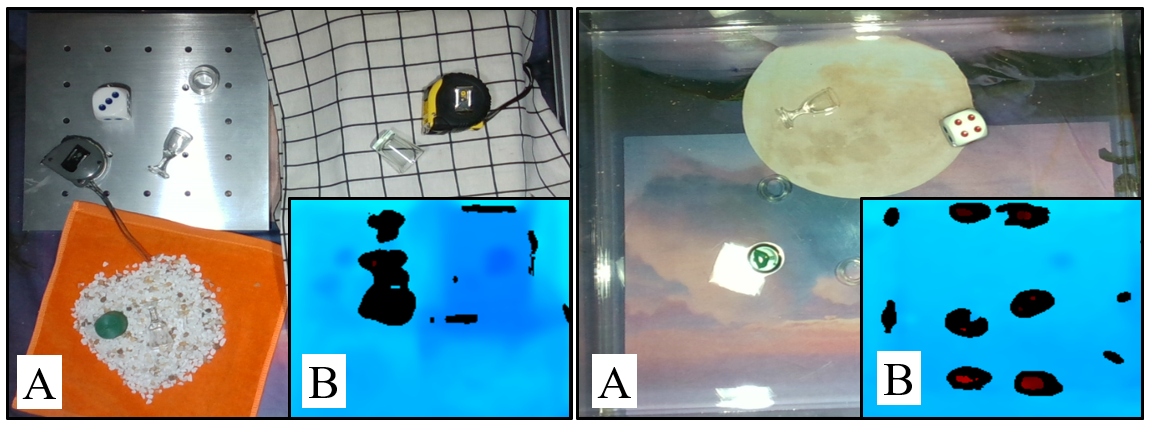}
	\caption{Detection with RGB and depth cameras. Left: undulating scenes: RGB (A) and Depth (B) images; Right: underwater scenes: RGB (A) and Depth (B) images. } \label{fig:depth}
\end{figure}

Secondly, object grasping on undulating planes is difficult with RGB cameras because it is hard to estimate the height where the object is placed. As shown in Fig.~\ref{fig:depth}, even by incorporating depth cameras, such a problem cannot be well solved for transparent object grasping. The reason is mainly bifold: on the one hand, the depth information for transparent objects is inaccurate, and on the other hand, undulating scenes have some shadows, overlaps, and reflective areas, which raises more challenges for transparent object detection. Therefore, the lack of accurate information about transparent objects and the interference of the environmental background is a difficult problem to solve, which is one of the reasons why we use RGB images rather than depth images to achieve transparent grasping position detection in complex backgrounds. Sand is a special undulating scene. In addition to the above problems, its surface is more uneven, where sand particles of different colors will also influence the detection, and sand is also easy to slide during the grasping process. To our knowledge, there is still a lack of studies or experiments for transparent object grasping on undulating scenes or sand environments.

\begin{table}[]
\caption{Related work on transparent object grasping}
\begin{center}
\begin{tabular}{m{2cm}<{\centering}|m{1cm}<{\centering}|m{4.2cm}<{\centering}}
\hline
Scenes & Difficulty & Related Work\\\hline
Plane      &   \multirow{1}{*}{Normal}        &   Weng \textit{et al.}~\cite{weng2020multi}, Sajjan \textit{et al.}~\cite{sajjan2020clear},\newline Liu \textit{et al.}~\cite{liu2020keypose}, Ichnowski \textit{et al.}~\cite{ichnowski2021dex},\newline Kerr \textit{et al.}~\cite{kerrevo}, Cao \textit{et al.}~\cite{cao2021fuzzy},\newline Jiang \textit{et al.}~\cite{jiang2022a4t}, Zhou \textit{et al.}~\cite{zhou2020lit}
\\\cline{1-3}
Fragments grasping   &  \multirow{3}{*}{Medium}     & N/A  \\

Stacking \& Overlapping      &         &  Zhou \textit{et al.}~\cite{zhou2019glassloc}, Lysenkov \textit{et al.}~\cite{lysenkov2013pose} \\ \cline{1-3}
Undulating     &  \multirow{3}{*}{High}     &       N/A    \\
Sand    &             &     N/A    \\ 
Underwater      &              &  Oberlin \textit{et al.}~\cite{oberlin2017time}, Zhou \textit{et al.}~\cite{zhou2018plenoptic} \\ \cline{1-3}
 Highly dynamic underwater &    \multirow{1}{*}{Very high}         &   N/A     \\ \hline 
\end{tabular}
\end{center}
	\label{tabt}
\end{table}
\begin{figure*}
	\centering
	\includegraphics[width=0.96\textwidth]{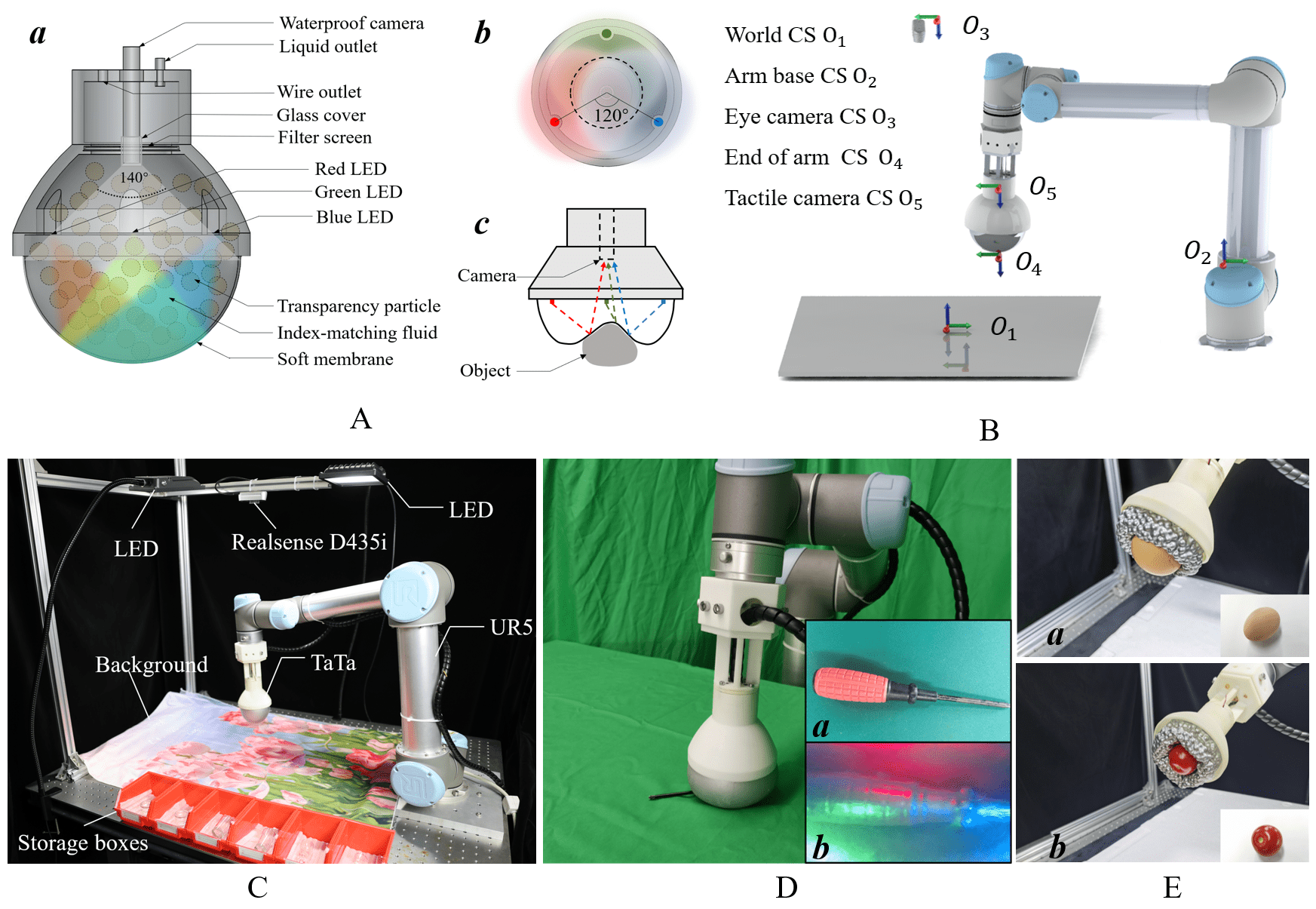}
	\caption{Hardware system. (A)  The structure of TaTa: (a) The schematic diagram of TaTa, (b) The layout of the inside LEDs, (c) The illustration of the inside light path.  (B) Coordinate system (CS). (C) Visual-tactile fusion grasping experimental platform. (D) Tactile perception effect test: (a) Screwdriver picture, (b) Perception result.  (E) Grasping performance testing: (a) Grasp an egg, (b) Grasp a tomato.   } \label{fig:Hardware}
\end{figure*}

Thirdly, transparent object grasping in underwater scenes is also challenging due to the similar optical properties of water and transparent objects. As shown in Fig.~\ref{fig:depth}, even with a depth camera, transparent objects are still undetectable in water, and there are many reflections on the surface of the water under the illumination of light, making things worse. To solve the problem of underwater transparent object detection, Zhou \textit{et al.}  proposed the Plenoptic Monte Carlo Localization (PMCL) method for localizing the pose of a translucent object underwater using a Lytro first-generation light field camera ~\cite{zhou2018plenoptic}.
Similarly, Oberlin \textit{et al.}  proposed a formal model for robotic light-field photography~\cite{oberlin2017time}, which can  turn a calibrated eye-in-hand camera into a time-lapse light-field camera. Although this method can be deployed on conventional cameras, thousands of RGB images from different angles need to be captured for one detection. Furthermore, these methods have not been studied for some dynamic underwater scenes with bubbles, waves, reflections, and complex backgrounds, which are extremely difficult, and even using a light field camera may probably fail.

In summary, most existing studies focus on grasping transparent objects with known shape in simple scenes such as on a plane, while several difficult scenes listed in Table~\ref{tabt} are rarely studied and still remain an open problem. 

\section{Hardware Setup}

The human hand has sensitive tactile perception because its surface is covered with dense tactile nerves~\cite{dargahi2004human}. Similarly, various tactile sensors have been designed for robots, such as piezoelectric~\cite{zhang2022finger}, capacitive~\cite{an2018transparent}, triboelectric~\cite{SONG2022106798}, and piezoresistive sensors~\cite{stassi2014flexible}, but the resolution is still not comparable to human hands. Thanks to the commercialization and miniaturization of the CMOS image sensors, a series of tactile detection devices based on optical imaging are invented and realize high-resolution sensing with low costs, e.g., GelSight~\cite{yuan2017gelsight}, GelSlim~\cite{donlon2018gelslim}, and FingerVision~\cite{trueeb2020towards}. However, such devices are mainly designed for fingertips and cannot acquire the overall contour of the contacted object. In addition, they usually adopted a silicone plus transparent acrylic sheet solution which has limited deformation capability.

To realize transparent object grasping, a universal soft gripper named TaTa is adopted here, as shown in Fig.~\ref{fig:Hardware}(A)(D), which has tactile perception on a large  hemispherical surface. Details of the TaTa gripper can be found in our previous paper~\cite{9811806}. Meanwhile, we upgrade the previous version of the TaTa gripper by using the camera with a larger imaging range and improving the waterproof ability to achieve better detection performance and durability. TaTa adopts the grasping principle of particle jamming and vision-based tactile detection technology, using the principle of refractive index matching to design a special solid-liquid mixture that looks totally transparent~\cite{goodman2015statistical}, overcoming the interference of internal particles on the internal camera. Hence, it has large-area, high-quality tactile detection ability as well as adaptive grasping ability. 

The hardware setup is depicted in Fig.~\ref{fig:Hardware}(B)(C). A RealSense D435i camera is fixed on the top frame as the “eye”, which can acquire $480\times640$ image information and the TaTa gripper is attached to the UR5 robotic arm. Two LEDs are used to provide lighting to the platform. We divide the system into five coordinate systems and use the center of the gripping plane as the origin of the world coordinate system $O_{1}$.
Firstly, we calibrate the intrinsics and extrinsics of the eye camera with a checkerboard~\cite{fetic2012procedure} to establish the relationship between the camera coordinate system $O_{3}$ and the world coordinate system $O_{1}$.
Secondly, since the robot arm and the gripping plane are at the same height, the relationship between the arm base coordinate system $O_{2}$ and the world coordinate system $O_{1}$ is derived by coordinate transformation.
Thirdly, the position of the end of the gripper in $O_{2}$ can be obtained through the official program interface of the UR5 robot arm so that the arm can be controlled to reach the location of the transparent object captured by the eye camera.
Finally, taking the gripper center as the origin and establishing the relationship between the tactile camera coordinate system $O_{5}$ of the tactile sensor and the coordinate system $O_{4}$ of the gripper end, so we can get the offset of the position where the contact between the gripper and the object occurs relative to the origin of the gripper, after which the offset is mapped to the displacement of the end of the arm to achieve tactile calibration.

To verify the capability of handling fragile objects, tests on grasping an egg and a tomato with TaTa are conducted, as shown in Fig.\ref{fig:Hardware}(E). Meanwhile, we upgrade the problem of the small imaging range of the previous version of the TaTa gripper by using the camera with a larger imaging range and improving the waterproof ability to achieve better detection performance and durability.

\section{Methodology}

\begin{figure*}
	\centering
	\includegraphics[width=1\textwidth]{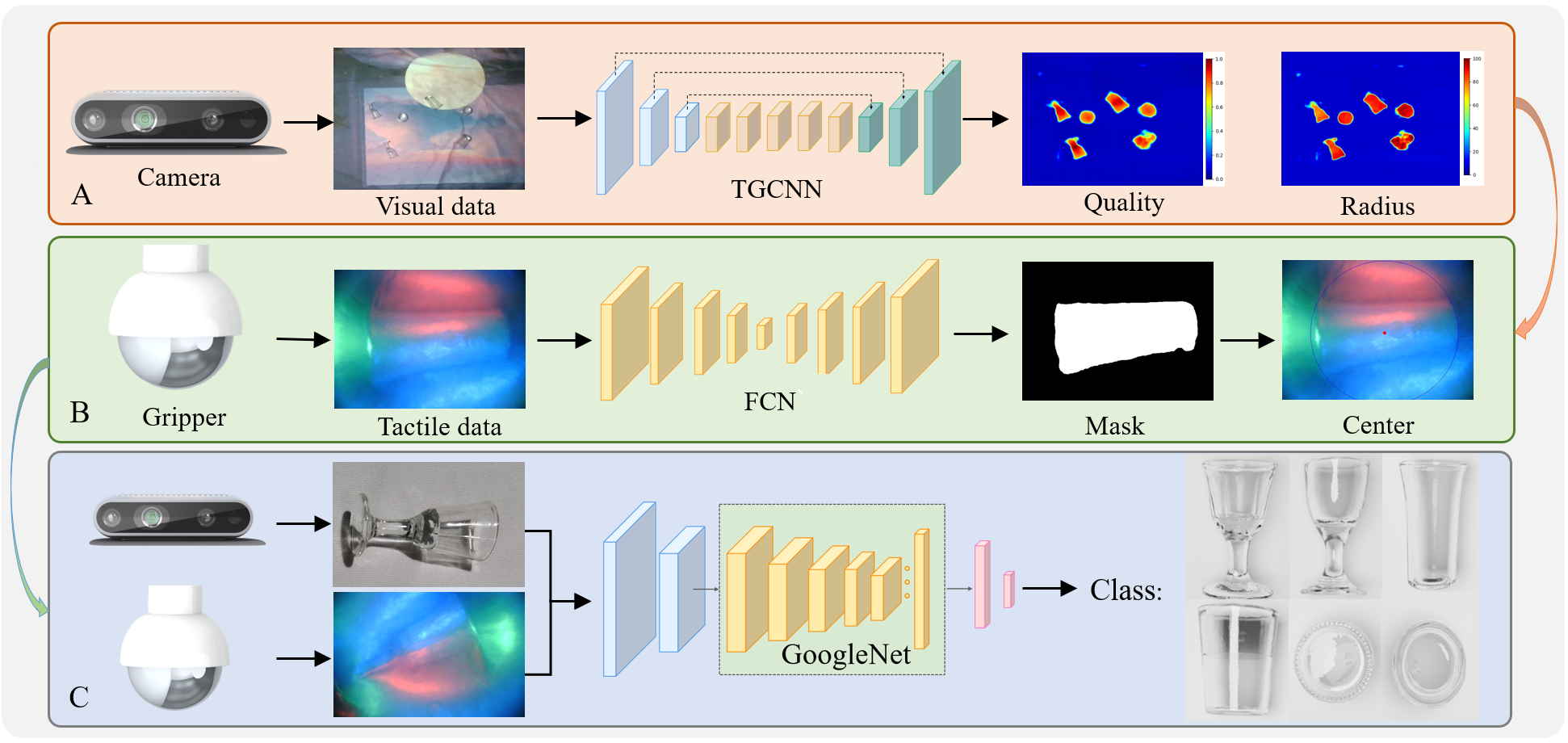}
	\caption{The visual-tactile fusion framework for transparent object grasping. (A) Grasping position detection. (B) Tactile information extraction.
 (C) Visual-tactile fusion classification.} \label{fig:framework}
\end{figure*}

This section introduces the algorithms used in our proposed visual-tactile fusion grasping framework. As shown in Fig.~\ref{fig:framework}, to achieve transparent object grasping, we propose a transparent object grasping position detection algorithm, a tactile information extraction algorithm, and a visual-tactile fusion classification algorithm, respectively. Besides, a Gaussian-Mask annotation method is also developed for our synthesized transparent object dataset.

\subsection{Dataset Generation and Annotation}
\begin{figure*}
	\centering
	\includegraphics[width=1\textwidth]{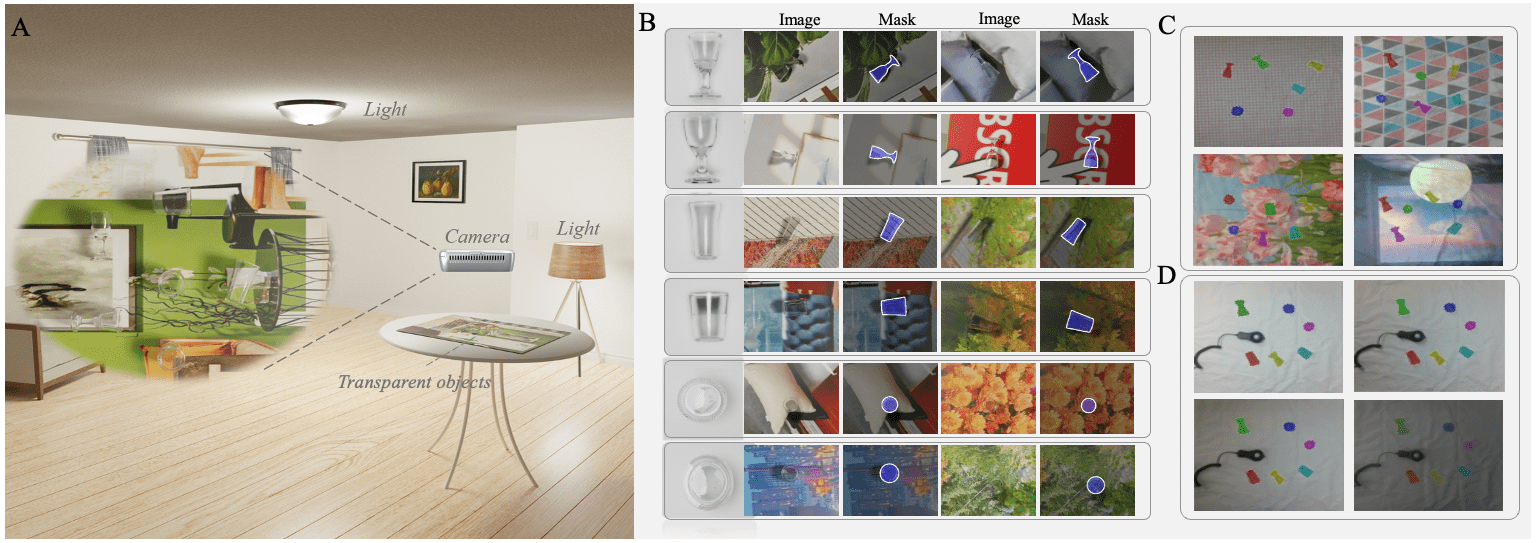}
	\caption{SimTrans12K Dataset. (A) Scene setup for generating the transparent object synthetic dataset using Blender. (B) Synthetic dataset of transparent objects in different backgrounds. (C) The real dataset in different backgrounds. (D) The real dataset in different brightnesses.} \label{fig:SimTrans12K}
\end{figure*}

 The neural network-based grasping position detection method requires a large dataset, and hence it is challenging to collect and annotate datasets manually. 
To tackle this problem, we adopt Blender to make a multi-background transparent object grasping dataset, SimTrans12K, which contains 12,000 synthetic images and 160 real images, as illustrated in Fig.~\ref{fig:SimTrans12K}. 
The reason to choose Blender is due to its high flexibility and capability to simulate the key features of transparent objects, such as surface reflections, refraction, and soft shadows.

SimTrans12K contains 6 types of objects and 2,000 different scenes. To obtain sufficiently complex and adequate backgrounds, we cropped some images from videos containing rich home decoration layouts and landscapes as backgrounds. The scene setup for generating the transparent object synthetic dataset is shown in Fig.~\ref{fig:SimTrans12K}(A). We use Blender 2.90's physically-based Cycles renderer with path tracing set to 256 samples pixel, and max light path bounces set to 1024. For glass materials, we set the index of refraction to 1.45 to match the physical glass. In each scene, two light sources are used to illuminate the location of the object and generate reflection spots on the surface of the object. The maximum power of the light is 1000 W and the minimum power is 100 W. A camera is placed above the transparent object, and the acute angle between the camera’s optical axis and the $z$-axis of the world coordinate is varied in the range $[0,\pi/24]$. Camera intrinsics are set the same as the RealSense D435i camera.

Based on the information obtained during rendering, ground truth labels are generated for training. Generative models such as the Generative Grasping CNN (GGCNN)~\cite{morrison2018closing} rely on the same binary ground truth generation during training. However, binary ground truth labels treat the edge and the center of annotation with the same weight, which is easy to make the grasping position deviate from the optimal center and come to the edge of the object. It can lead to a declined grasping success rate and even damage the object. To improve the reliability of dataset annotation, we propose a transparent object grasping position annotation method based on Gaussian distribution and the transparent object mask (Gaussian-Mask). Previously, Hu \textit{et al.} proposed the idea of using Gaussian distribution for object grasping position annotation. A Gaussian distribution rectangular box was adopted for the annotation of the gripping position, which works well for ordinary objects~\cite{cao2021lightweight}. However, the grasping position detection of transparent objects is much more challenging since the texture of transparent objects changes with the backgrounds. Fortunately, although the texture properties of transparent objects may change dramatically, their boundary information is relatively stable~\cite{xie2020segmenting}. Therefore, instead of using rectangular box annotation, we directly use the mask of the transparent object itself as the grasping contour and use Gaussian distribution to represent the optimal grasping position, which makes full use of the boundary information of the transparent object.

For the TaTa gripper, which achieves gripping by wrapping the whole object, we can use the center of the transparent object as the optimal grasping position. To make the grasping position as close as possible to the object center, a grasping quality distribution map that satisfies a Gaussian distribution from the center of the annotated object is generated. In this way, the point near the object center has a higher grasping quality than the point away from the center, so it is easier for the gripper to select the object center for grasping. Positioning in the object center helps TaTa for better tactile detection and also reduces the probability of potential damage to the object during grasping. To make the Gaussian-Mask annotation adapt to different objects and camera positions in the scene, the two farthest points on the object on the same $x-y$ plane are selected. Then the two points are projected onto the rendered image. The center of the line connecting the two points is chosen as the center of the Gaussian distribution, and half the distance between the two points is determined as the Gaussian distribution radius. However, the Gaussian-Mask annotation method also has some limitations. This method uses the object's mask as the annotation frame and is more suitable for objects whose center is located on the object.
\begin{figure}
	\centering
	\includegraphics[width=0.46\textwidth]{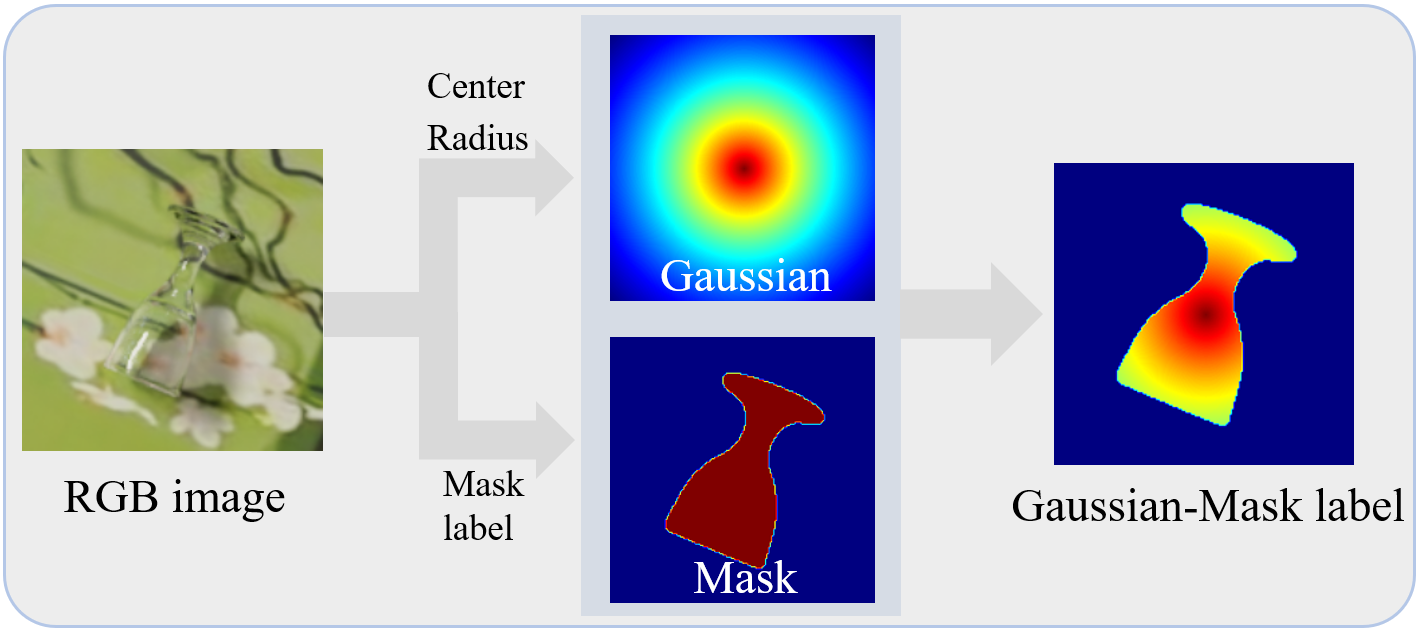}
	\caption{Gaussian-Mask annotation process of transparent objects.} \label{fig:gaussian}
\end{figure}
The Gaussian-Mask annotation process is shown in Fig.~\ref{fig:gaussian}. The object center, Gaussian distribution radius, and Gaussian-Mask labels are obtained directly from the RGB image, which is further processed to obtain the Gaussian representation ground truth labels. Thanks to the Gaussian-Mask annotation, the grasping network could regress to a more accurate grasping center.

\begin{figure*}
	\centering
	\includegraphics[width=0.94\textwidth]{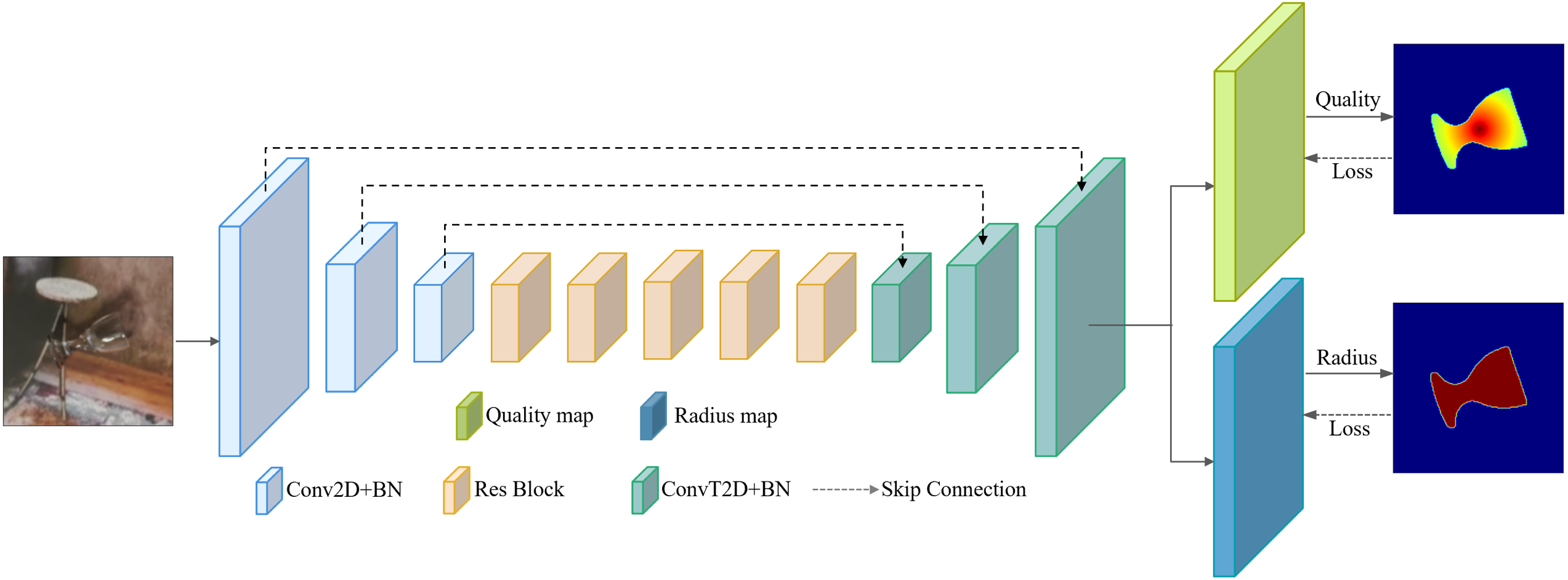}
	\caption{The network architecture of the proposed TGCNN.} \label{fig:TGCNN}
\end{figure*}

\subsection{Transparent Object Grasping Position Detection}

Due to the unique optical properties, the appearance of transparent objects is easily disturbed by the backgrounds. In many cases, it is difficult for even humans to directly distinguish the types of transparent objects under different backgrounds, which makes the classification and grasping of transparent objects by vision difficult. Therefore, we adopt the generative grasping model, which can directly generate the grasping positions from images without recognition and classification. The model is expected to learn the general properties of transparent objects and then apply them to detect and grasp unseen transparent objects in complex backgrounds.

\textbf{(a) Network Architecture:} Fig.~\ref{fig:TGCNN} shows the proposed TGCNN model, which is a generative architecture taking in a 3-channel RGB image and generating pixel-wise grasps in the form of two images. The 3-channel RGB image is passed through convolutional layers, residual layers, and convolution transpose layers to generate two images. Each residual layer contains two convolutional layers, two batch normalization layers, and a shortcut connection. At the same time, the skip connections in the network enable the network to obtain more hierarchical information fusion, which makes the network more effectively combine scene information to detect transparent objects. TGCNN  has 2,145,154 parameters with a trained model size of 8.23 MB.
The output image of the network includes grasping quality and grasping radius, and these two parameters can guide the grasping for the TaTa gripper.

\textbf{(b) Grasp Definition:} A grasp perpendicular to the $x$-$y$ plane is defined as $\mathbf{g}_{r} = (\mathbf{p},r)$. The grasp is described by the projection of the center position of the gripper $\mathbf{p}:(x,y)$ onto the $x$-$y$ plane and the height $h$ between the gripper center and the $x$-$y$ plane in Cartesian coordinates. Since the shape of the gripper is hemispherical, the flexibility of the gripper enables it to be deformed. 
For the same gripper and object, the lower the gripper center from the $x-y$ plane after contact with the object, the more the gripper is compressed and the larger the contact area between the gripper and the object will be. 
 
\begin{figure}
	\centering
	\includegraphics[width=0.48\textwidth]{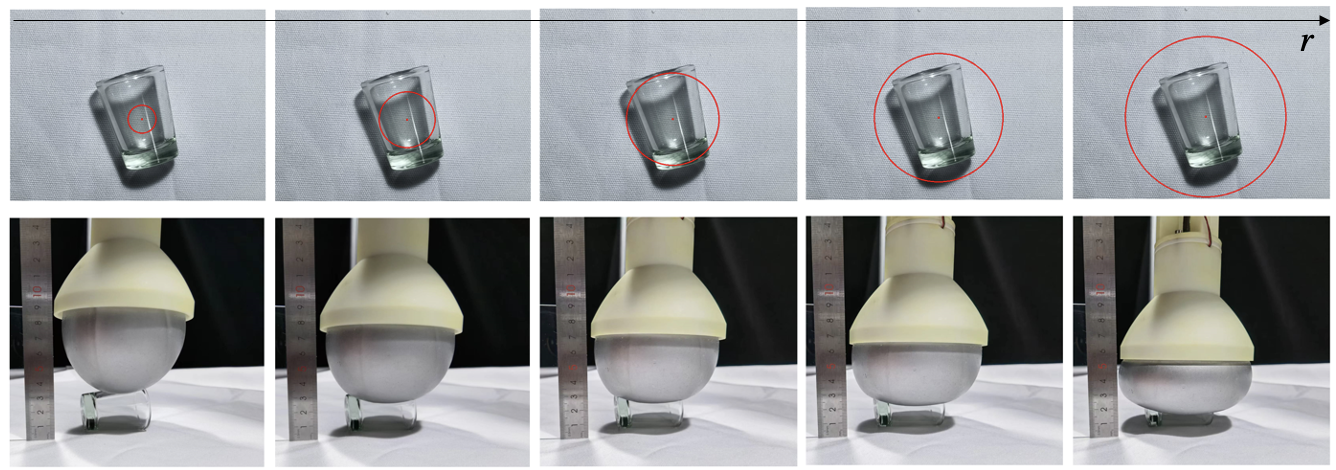}
	\caption{The influence of different grasping radius $r$ on detection.} \label{fig:radius}
\end{figure}

Since the contact surface is similar to a circle, the index grasping radius $r$ was used to describe the contact area sizes caused by different heights $h$ of the gripper. The influence of different grasping radius $r$ on detection is shown in Fig.~\ref{fig:radius}. A scalar quality measure $q$, representing the chances of grasp success, is added to the pose. To further improve the grasping efficiency of the gripper, an adaptive height-dropping method (AHD) is proposed. AHD can determine the distance between the gripper and the detection surface according to the size of the object. We use a small drop height for small objects and use a larger drop height for large objects. Because for small objects, a smaller drop height can make the gripper obtain complete tactile detection information.  While for a large object, a larger drop height can obtain a more complete tactile image of the object. The generated grasping radius is smaller when the object detection uncertainty is large, which ensures a balance between safety and detection efficiency when detecting in highly uncertain environments.

Assume we want to detect grasps through an RGB image $\mathbf{P} \in {\mathbb{R}^{m \times n \times 3}}$, with known camera intrinsic parameters. In the image coordinate system, a grasp is described by
\begin{equation}
\mathbf{g}_{i} = (\mathbf{s},r_{i},q),
\label{eqn1}
\end{equation}
where $\mathbf{s} = (u,v)$ is the centre and $r_{i}$ is the grasping radius in image coordinates.  In order to perform the grasp in image space on the robot, we can convert the image coordinates to the robot's frame of reference by the following transformation:
\begin{equation}
\mathbf{g}_{r} = {\xi_{RC}}({\xi_{CI}}(\mathbf{g}_{i})),
\label{eqn2}
\end{equation}
where  ${\xi_{RC}}$ is a transformation from the camera frame to the world frame and ${\xi_{CI}}$ is a transformation from 2D image coordinates to the 3D camera frame.

The above notation can represent multiple grasps in an image. The collective group of all grasps can be denoted as
\begin{equation}
\mathbf{G}=(\mathbf{R}, \mathbf{Q}) \in \mathbb{R}^{m \times n \times 2},
\label{eqn3}
\end{equation}
where $\mathbf{R}$ and $\mathbf{Q}$ $\in \mathbb{R}^{m \times n}$  contain values of grasping radius $r_{i}$ and quality measure $q$ respectively at each pixel $\mathbf{s}$.

Grasp candidates $\mathbf{g}_{i}$ are wanted to create directly by calculating the RGB images, so a mapping $\phi$ from RGB images to grasp map in the image coordinates was defined: $\phi(\mathbf{P}) = \mathbf{G}$. From $\mathbf{G}$ the best visible grasp in the image space ${\mathbf{g}_{i}^{*}} = \mathop {\max }\limits_\mathbf{Q} \mathbf{G}$ can be calculated, and the equivalent best grasp in world coordinates ${\mathbf{g}_{r}^{*}}$ can be obtained as well. For the case of multiple objects in the same scene, we will sort the visible grasps according to the quality measure $q$, and select the first $k$ visible grasps ($k$ is manually specified).

Huber loss is used for network training, written as:
\begin{equation}
\mathcal{L}\left(\mathbf{G}_{i}, \hat{\mathbf{G}}_{i}\right)= \begin{cases}0.5(\Vert\mathbf{G}_{i}-\hat{\mathbf{G}}_{i}\Vert_\text{F})^{2}, & \text { if }\Vert\mathbf{G}_{i}-\hat{\mathbf{G}}_{i}\Vert_1<1 \\ \Vert\mathbf{G}_{i}-\hat{\mathbf{G}}_{i}\Vert_{1,1}-0.5, & \text { otherwise }\end{cases}
\label{eqn4}
\end{equation}
Here, $\mathbf{G}_{i}$ denotes the grasp candidate which can be generated by the network, and $\hat{\mathbf{G}}_{i}$ is the ground truth grasp. 
$\Vert \cdot \Vert_\text{F}$ and $\Vert \cdot \Vert_{1,1}$ represent the Frobenius and “Entry-wise” L1 matrix norms, respectively.


\subsection{Tactile Information Extraction Algorithm}
\begin{figure*}
	\centering
	\includegraphics[width=0.96\textwidth]{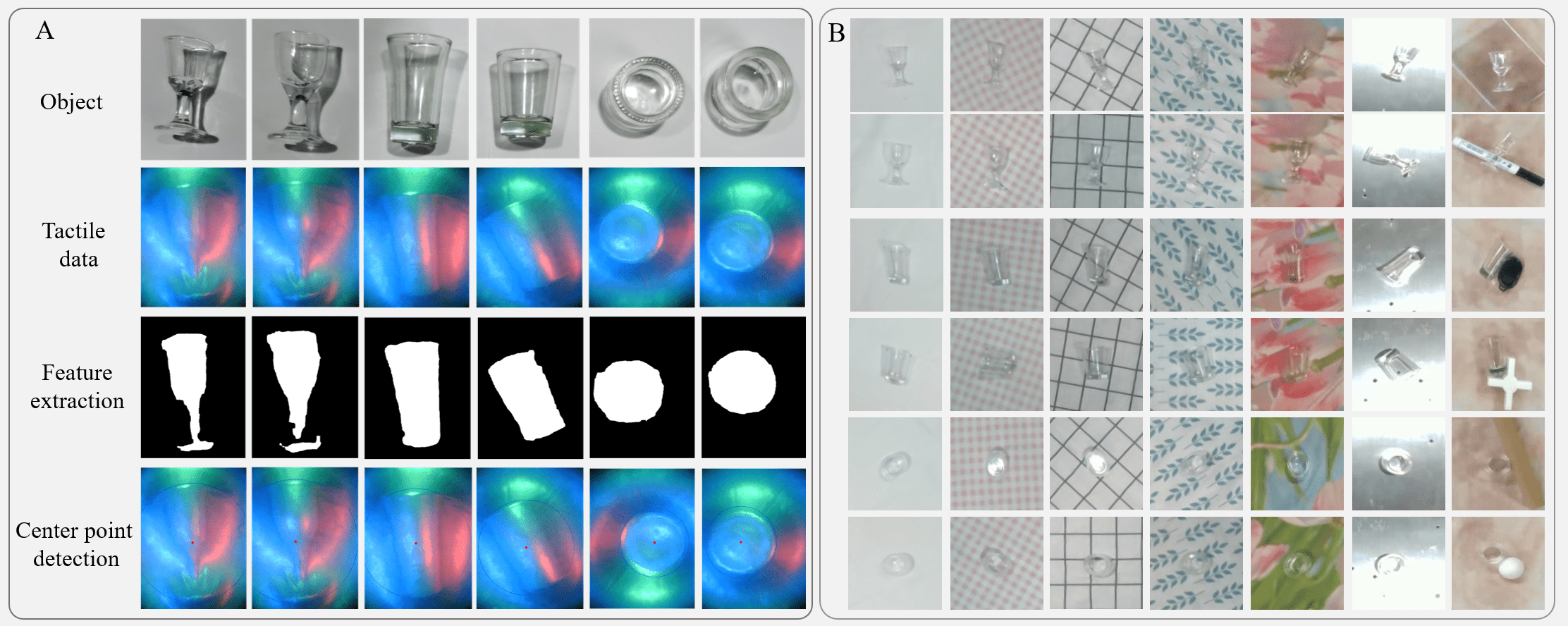}
	\caption{Visual-tactile fusion classification dataset. (A) Tactile data, FCN feature extraction, and center point detection results. (B) Visual classification dataset. } \label{fig:tactile_RGB}
\end{figure*}

To obtain the contact information between the gripper and the object, we extract the contour of the contact area using Fully Convolutional Networks (FCN). Compared with the frame difference method\cite{singla2014motion} and optical flow method\cite{brox2009large}, the FCN-based tactile feature extraction algorithm has stronger robustness and can still obtain clear contact information even if the internal optics of the sensor is changed. We acquired 160 images of the object in contact with the gripper as a training set and annotated the data at the pixel level, and some results are illustrated in Fig.~\ref{fig:tactile_RGB}(A). After 60 rounds of training, the segmentation accuracy achieves 98\%.

\begin{figure}[!t]
	\centering
	\includegraphics[width=0.48\textwidth]{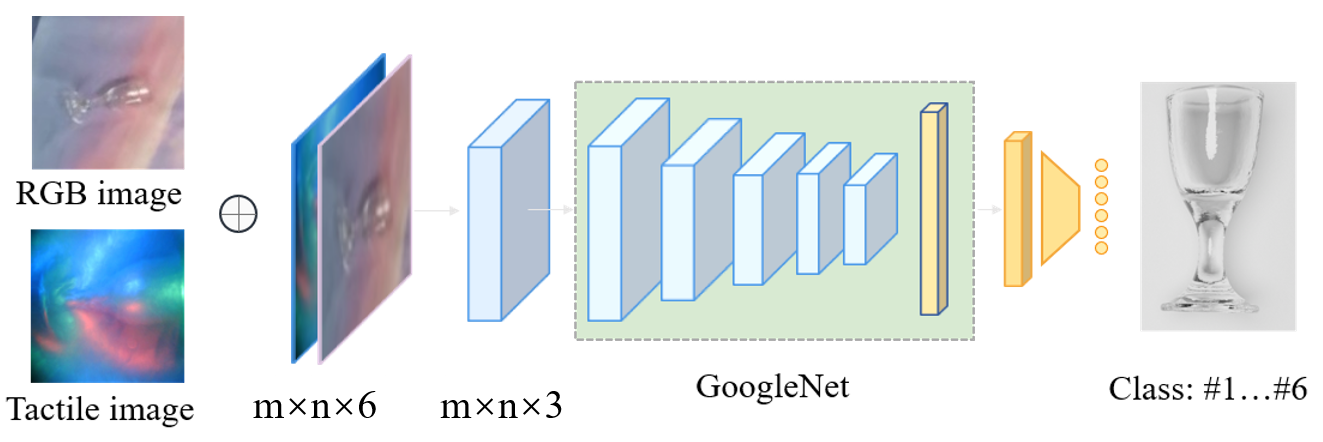}
	\caption{Visual-tactile fusion classification framework for transparent objects.} \label{fig:10}
\end{figure}

\subsection{Visual-tactile Fusion Classification}

Transparent objects have little visual information and the surface pattern changes with the backgrounds as well as the lighting conditions, making it difficult to classify by vision only. To solve this problem, a vision-tactile fusion method for transparent object classification is proposed, where RGB and tactile images are concatenated together for classification using GoogleNet, as depicted in Fig.~\ref{fig:10}. We collected 1,200 data of 6 objects with different backgrounds such as reflections, patterns, colors, overlapping, and stacking scenes as the training set and 600 data as the test set, as shown in Fig.~\ref{fig:tactile_RGB}(B).

To test the performance of the algorithm, we compare the visual classification and visual-tactile fusion classification algorithms. The visual classification accuracy is 59.3\%, while the visual-tactile fusion classification accuracy reaches 98.4\%, which increases the classification success rate by 39.1\%.

\section{Grasping Strategy}

\begin{figure}[!t]
	\centering
	\includegraphics[width=0.48\textwidth]{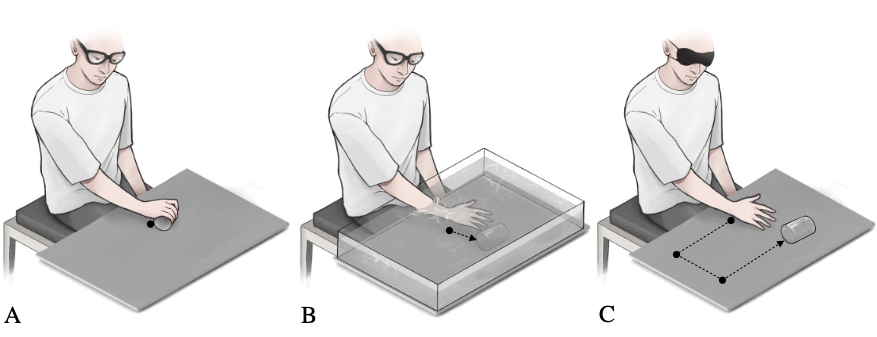}
	\caption{Human grasping strategies in different scenes (The black curve indicates the movement path of the hand). (A) Grasping objects in clear view. (B) Grasping transparent objects underwater. (C) Grasping objects in visually undetectable scenes. } \label{fig:human}
\end{figure}

Based on the algorithms proposed in the previous section, this section explains how to integrate them to accomplish transparent object grasping in different scenes, which forms the high-level grasping strategy for our visual-tactile fusion framework. We decompose a grasping task into three sub-tasks, i.e., object classification, grasping position determination, and grasping height determination. Each sub-task can be conducted by vision, touch, or fusion. 
Similar to humans, when vision can directly obtain the precise position of the object, we can control the hand to directly reach the object and complete the grasp, as shown in Fig.~\ref{fig:human}(A). When the vision can not accurately obtain the object's position information, we will use the tactile perception of the hand to slowly adjust the grasping position after obtaining the object's general position information until it touches the object and reaches the appropriate grasping position, as shown in Fig.~\ref{fig:human}(B). For object grasping in visually limited situations, as shown in Fig.~\ref{fig:human}(C), we will use the hand's rich tactile nerve to search for the position of the object in a wide range, which obviously wastes more time but is an effective way to solve the object grasping in these special scenes.

\begin{figure}
	\centering
	\includegraphics[width=0.48\textwidth]{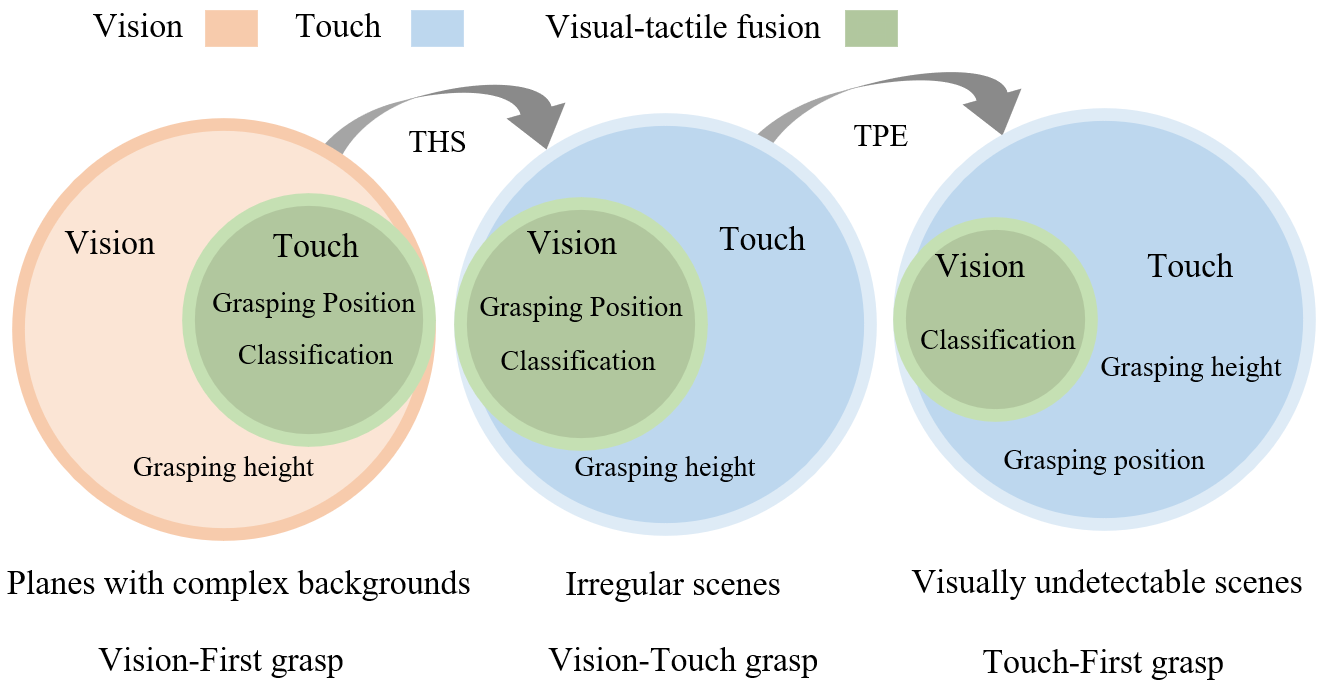}
	\caption{Grasping strategies in different scenes. The orange, blue, and green colors represent the functions of visual detection, tactile, and visual-tactile fusion, respectively. The framework can be adapted to different scenes by adjusting the grasping strategies.
 } \label{fig:fa}
\end{figure}

Inspired by human grasping strategies, we divide transparent object-grasping tasks into three types: planes with complex backgrounds, irregular scenes, and visually undetectable scenes, as shown in Fig.~\ref{fig:fa}. In the first type where vision is very effective and plays a key role, we use vision-first grasp. In the second type where vision and touch can work synergistically, we use vision-tactile grasp. While 
in the last type where vision may fail and touch becomes dominant in the task, we use touch-first grasp. Details of the three grasping strategies are introduced below.

\begin{figure}
	\centering
	\includegraphics[width=0.48\textwidth]{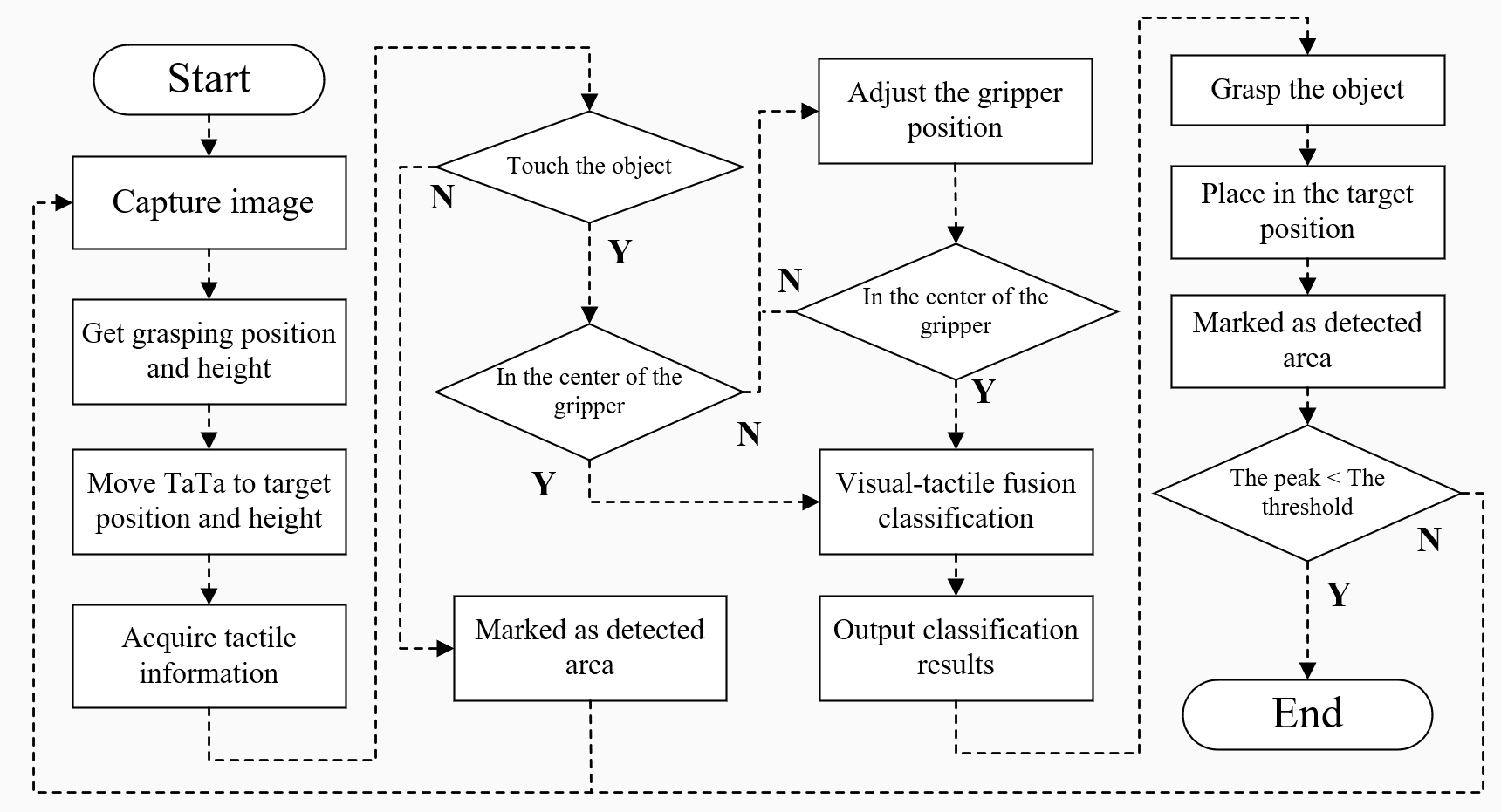}
	\caption{Flowchart of transparent object grasping on a plane with complex backgrounds.} \label{fig:f1}
\end{figure}

\subsection{Planes with Complex Backgrounds -- Vision-First Grasp}

Grasping objects on a plane can be achieved by visual detection~\cite{choi2018learning,mahler2017dex}, but the texture information of transparent objects changes with the background, so grasping transparent objects in complex backgrounds is challenging even on a plane. To tackle this, we propose a strategy for transparent object grasping with visual-tactile fusion, as depicted in Fig.~\ref{fig:f1}, which mainly includes three steps.

\begin{figure}
	\centering
	\includegraphics[width=0.48\textwidth]{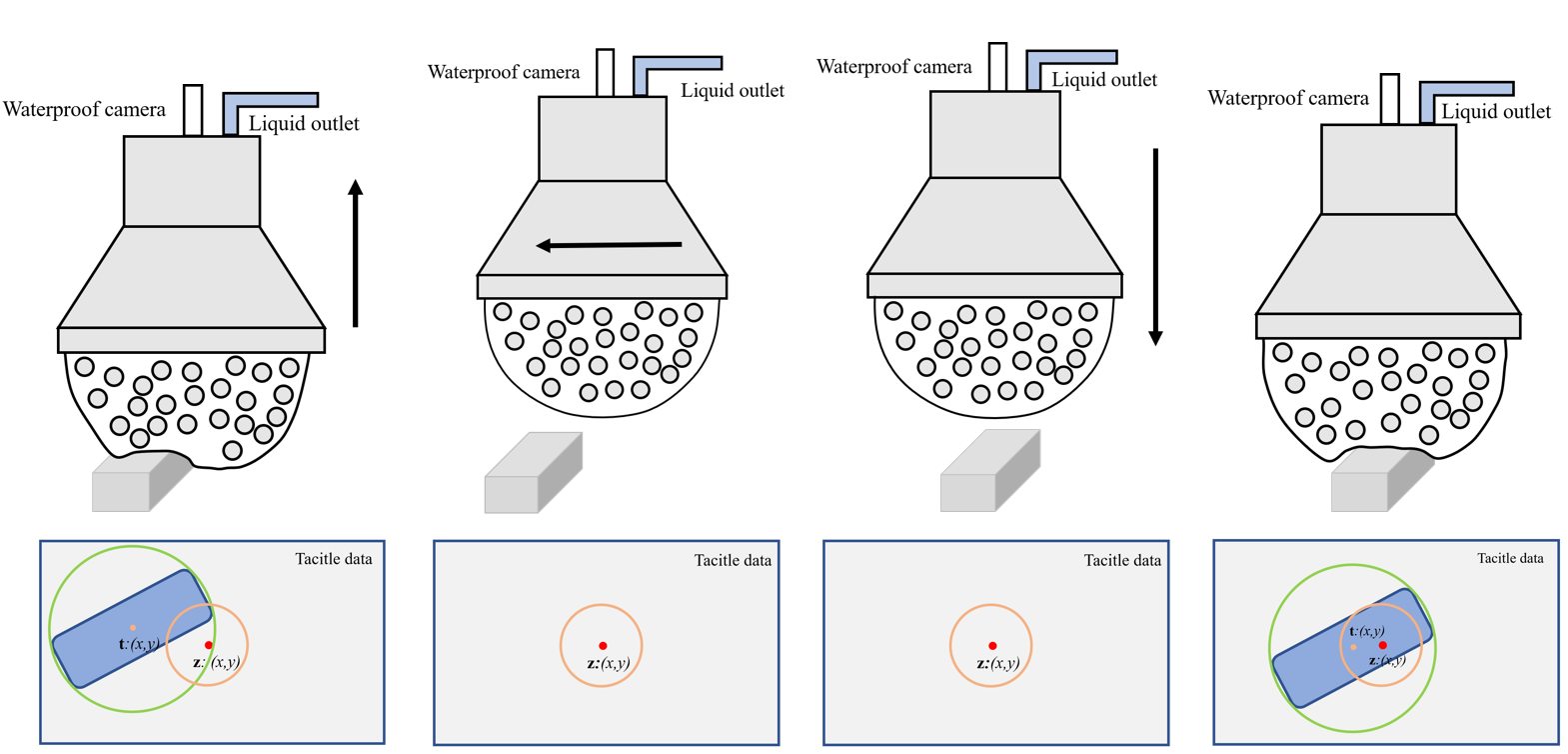}
	\caption{Tactile calibration grasping process.
	Firstly, the gripper touches the object, the center of the object outline does not appear in the gripper center, and the displacement between the object and the gripper is calculated as $\mathbf{d}:(x,y)$. Secondly, lift the gripper. Thirdly, move the gripper with a distance of $\mathbf{d}$. Fourthly, the gripper touches the object again. If the object center coincides with the gripper center, then calibration is completed.} \label{fig:7}
\end{figure}

\textbf{(a)} Use TGCNN to obtain the grasping position and height of the transparent object as the target position and control the gripper to reach the target position.
  
\textbf{(b)} Use tactile information to check if the gripper contacts the object. If not, mark it as a wrong detection point and proceed to the next position. If yes, the tactile information will be applied to further adjust the gripping position. Here, to get a more precise grasping position, we propose a tactile calibration method, as illustrated in Fig.~\ref{fig:7}. The method first uses a tactile information feature extraction network to segment the contact area~\cite{long2015fully}, which has been introduced in section IV. Then, use the minimum outer circle detection algorithm to obtain the circle center of the contact area, calculate the position relationship between the gripper center $\mathbf{z}:(x,y)$ and the minimum outer circle center $\mathbf{t}:(x,y)$ obtained by the tactile feature extraction $\mathbf{d} = \mathbf{z}-\mathbf{t}$, and control the gripper to move the distance of $\mathbf{d}$. During this stage, the tactile calibration algorithm will continue running until the object locates in the gripper center.

\textbf{(c)} Use the visual-tactile fusion framework to classify and place the object at a given place. Finally, the area is marked as detected and will not be revisited. The use of the grasping position marker prevents invalid revisiting especially when there are many interference areas in visual detection results.

\subsection{Irregular Scenes -- Vision-Touch Grasp}
\begin{figure}
	\centering
	\includegraphics[width=0.48\textwidth]{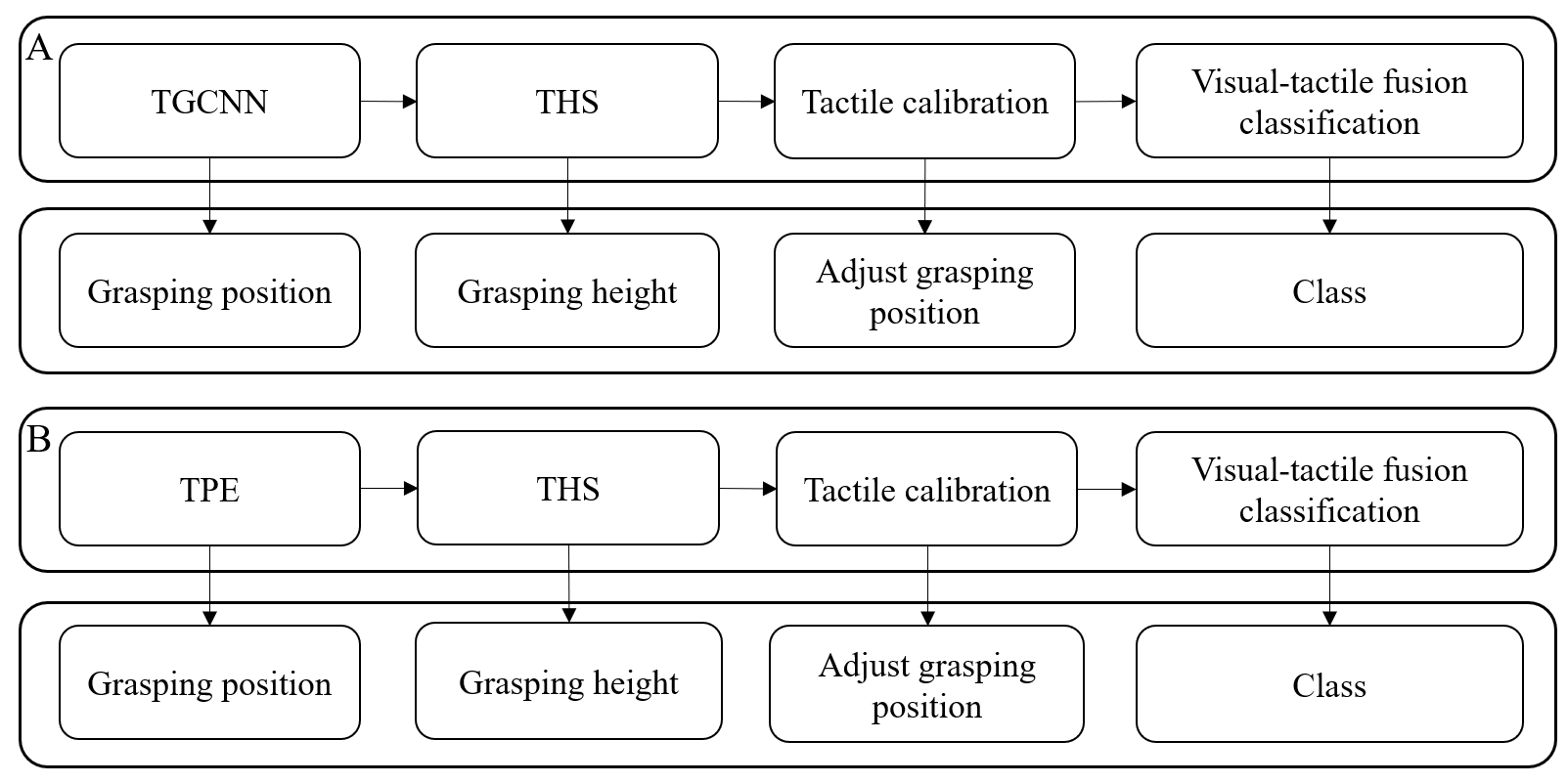}
	\caption{Transparent object grasping process: (A) In irregular scenes, (B) In visually undetectable scenes.} \label{fig:f2}
\end{figure}

Compared to grasping objects on planes, it is more challenging to grasp transparent objects in irregular scenes, such as overlapping, stacking, and undulating scenes. For stacking and overlapping scenes, it is difficult to separate two objects with similar textures by RGB vision detection either for transparent or non-transparent objects. And for undulating surfaces, it is difficult to obtain the precise grasping height of the object simply by using an RGB camera. 

So we add the THS module which uses tactile sensing to adjust the grasp height. The implementation process is shown in Fig.~\ref{fig:f2}(A). First, we still use TGCNN to get the grasping position of the transparent object, but the height of the object cannot be determined. Therefore, when the gripper reaches the specified position, the THS module will be activated. The gripper is controlled to keep exploring downward until it touches the object or reaches the lowest point, and then complete grasping the object through tactile calibration. The THS module not only enables the grasping of objects in undulating scenes but also solves the problem of grasping transparent objects in overlapping and stacking scenes.

\subsection{Visually Undetectable Scenes -- Touch-First Grasp}
Although vision is a powerful detection method, it may fail in some scenes, such as transparent object detection in highly dynamic underwater scenes. Because water and transparent objects have similar optical properties, the water flow and ripples will result in difficulties to detect the grasping position by vision. For transparent objects, we define scenes such as highly dynamic underwater, darkness, and smoke as visually undetectable scenes.

To achieve grasping in visually undetectable scenes, we add a TPE module as introduced in the previous section. The implementation process is shown in Fig.~\ref{fig:f2}(B). Firstly, it uses touch to search for the transparent object in a specific range. When in contact with the object, it uses THS to determine the grasping height and tactile calibration to determine the grasping position. Finally, it applies visual-tactile fusion for classification. Therefore, when vision is not effective, we can use touch to obtain both the grasping height and position, like human grasping in the dark. The advantage of this method is that object grasping can still be achieved even without vision, while the disadvantage is that it is inefficient and may fail to find the object when the exploration area is too large.

\section{Experiments}
This section presents the experimental results of the proposed algorithms and visual-tactile fusion grasping framework. Firstly, to test the effectiveness of our proposed transparent object dataset, the annotation method, and the grasping position detection network, we conduct synthetic data detection experiments (\textbf{Exp. 1}) and transparent object grasping position detection experiments under different backgrounds (\textbf{Exp. 2}) and brightness (\textbf{Exp. 3}). Secondly, to verify the effectiveness of the visual-tactile fusion grasping framework, transparent object classification grasping experiments (\textbf{Exp. 4}) and transparent fragment grasping experiments (\textbf{Exp. 5}) are designed. Thirdly, we design transparent object grasping experiments in irregular (\textbf{Exp. 6}) and visually undetectable scenes (\textbf{Exp. 7}) to test the effectiveness of the framework after adding the THS module and the TPE module. The failed trials and limitations are provided and discussed as well.

\subsection{Exp. 1: Object Detection with Synthetic Data}

To evaluate the performance of the TGCNN algorithm, we apply the index of grasping overlap degree (GOD) to measure if a detection is successful, similar to Saxena \textit{et al.}~\cite{saxena2008robotic} and Jiang \textit{et al.}~\cite{jiang2011efficient}. If a calculated grasp circle and the label mask 
share an intersection (i.e., the GOD) greater than 45\%, a detection is considered to be correct. 

Most of the current research on transparent object grasping such as ClearGrasp, Dex-NerF, and LIT is based on depth complementation and pose estimation of RGB-D information, which cannot be directly compared to  our algorithms that are based on RGB data.
To make a comparison, we consider the currently more mainstream generative grasp networks such as GGCNN~\cite{morrison2018closing}, Redmon~\cite{redmon2015real}, and GI-NNET~\cite{shukla2022generative}. Since most of these networks are designed for parallel two-finger grippers based on RGB-D input, 
modifications are needed to enable them to operate on our proposed dataset---we change the input data to RGB and change their output from the width and angle to  the radius.

In the experiments, we evaluate TGCNN from multiple aspects: (a) Image-wise evaluation with unseen backgrounds; (b) Object-wise evaluation with unseen objects; (c) Evaluation of Gaussian representation; (d) Multi-object evaluation in cluttered scenes.

\textbf{(a) Image-wise Evaluation in Unseen Backgrounds:} For transparent objects, changes in the background can greatly affect their visual features and may result in recognition errors and grasping failures. To evaluate the performance of TGCNN, we select 6 objects for training and then test the detection accuracy on unseen backgrounds. The training set contains 4,000 images and the testing set contains 1,000 images with different backgrounds from the training dataset. Although the proposed synthetic data rendering scheme can generate a large number of transparent object data easily, we hope that the network can learn with not too much data.

\begin{figure*}
	\centering
	\includegraphics[width=0.96\textwidth]{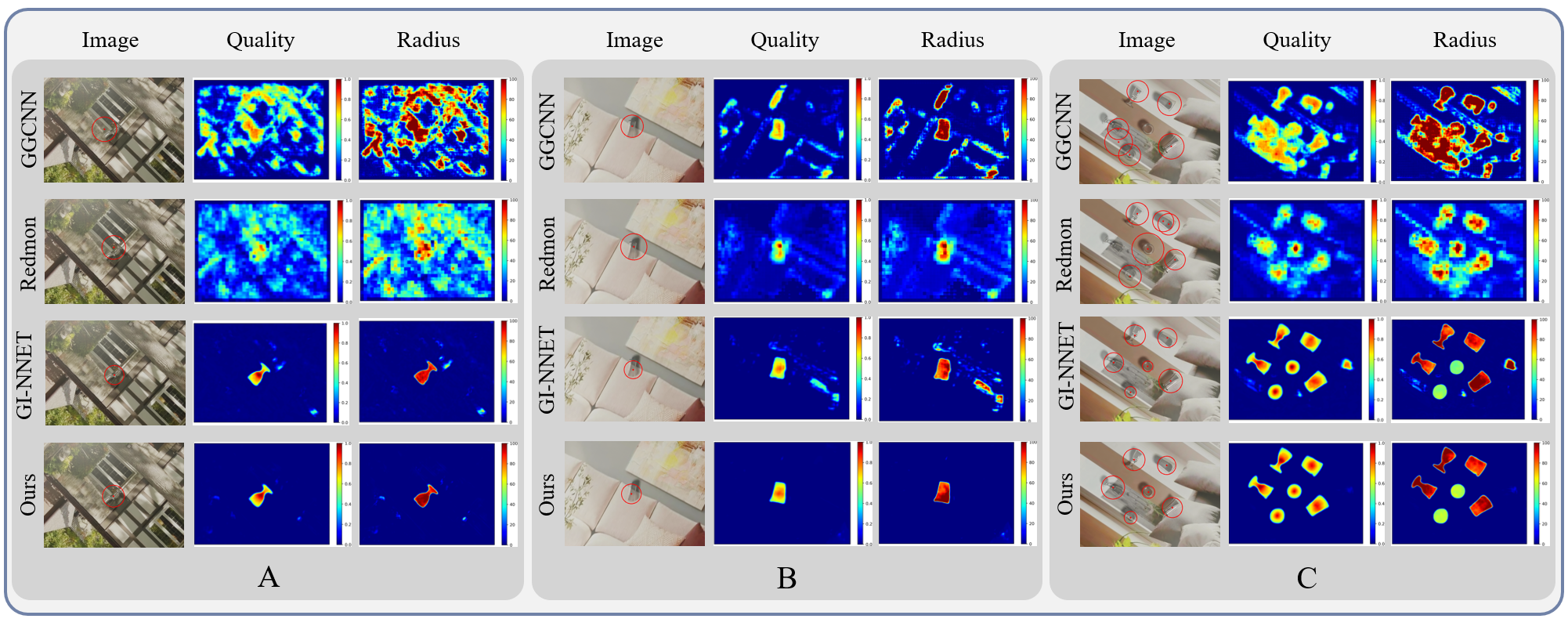}
	\caption{Synthetic dataset detection results. (A) Image-wise evaluation in unseen backgrounds. (B) Object-wise evaluation with unseen objects. (C) Multi-object evaluation in cluttered scenes.} \label{fig:synthetic}
\end{figure*}

As a result, TGCNN successfully detects a total of 942 objects within the test set, with an accuracy of 94.2\%. The results are compared with some currently well-known and open-source algorithms, as shown in Table~\ref{tab1}. Accuracy (Gaussian-Mask, \%) in Table~\ref{tab1} means we use the Gaussian representation of the label, and accuracy (Binary, \%) means we use the binary representation of the label. Fig.~\ref{fig:synthetic}(A) shows the detection results of each algorithm, indicating that our algorithm has better performance in unseen backgrounds.

\begin{table}[h]
	\centering	
	\caption{Detection results in unseen backgrounds}
	\begin{tabular}{m{1.8cm}<{\centering}|m{2.7cm}<{\centering}|m{2.6cm}<{\centering}}
		\hline
		Algorithm    & Accuracy (Gaussian-Mask)   &  Accuracy (Binary)\\ \hline 
		GGCNN \cite{morrison2018closing}    & 67.2\%     & 27.2\%  \\ 
		Redmon \cite{redmon2015real}  &  59.3\%    &  42.5\% \\ 
        GI-NNET \cite{shukla2022generative}   & 89.3\%     &  25.1\% \\
		\textbf{Ours}     &  94.2\% &     25.3\%  \\ \hline
	\end{tabular}
	\label{tab1}
\end{table}

\textbf{(b) Object-wise Evaluation with Unseen Object:} Besides the good performance under new backgrounds, TGCNN can also achieve grasping position detection for unseen objects. To test this, we use two objects from the dataset as the training set and the remaining four objects as the testing set. Furthermore, we also included four objects from the LIT dataset\cite{zhou2020lit} in the test set to further guarantee the generalization of the dataset. The training set contains 4,000 images and the testing set contains 1,000 images with unseen objects but the same backgrounds as the training set. As can be seen from Table~\ref{tab2}, TGCNN outperforms other algorithms in the detection accuracy of unseen objects. Fig.~\ref{fig:synthetic}(B) shows the detection results of TGCNN and other algorithms.

\begin{table}[h]
	\centering	
	\caption{Detection results with unseen objects}
	\begin{tabular}{m{1.8cm}<{\centering}|m{2.7cm}<{\centering}|m{2.6cm}<{\centering}}
		\hline
		Algorithm   & Accuracy (Gaussian-Mask) &Accuracy (Binary)   \\ \hline
		GGCNN~\cite{morrison2018closing}      & 83.8\% &  44.2\% \\    
		Redmon~\cite{redmon2015real}   &  48.6\%   &  30.1\% \\ 
            GI-NNET~\cite{shukla2022generative} &      96.1\%     &   61.2\%   \\
	\textbf{Ours}      &     99.2\%   & 70.9\%    \\ \hline
	\end{tabular}
	\label{tab2}
\end{table}

\begin{figure}
	\centering
	\includegraphics[width=0.48\textwidth]{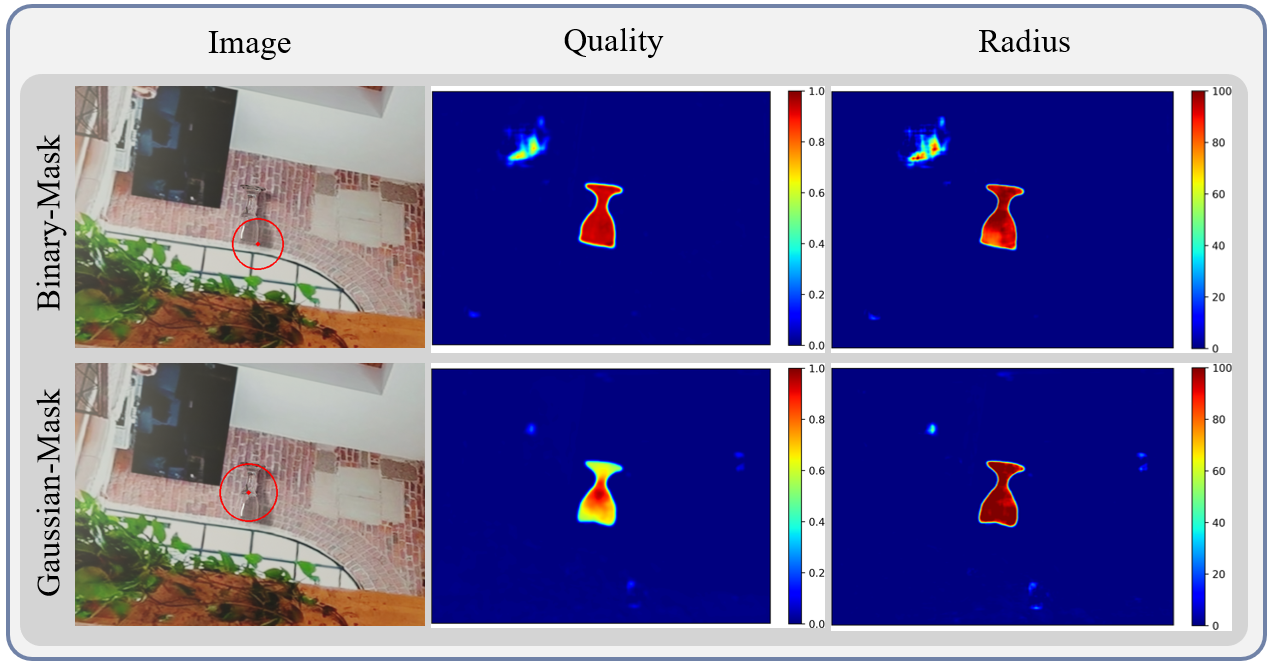}
	\caption{Detection comparison between Gaussian and binary representations.} \label{fig:effectcomparison}
\end{figure}

\textbf{(c) Evaluation of Gaussian Representation:} During the experiments, we found that the utilization of Gaussian representation in the annotation plays a key role in improving the detection accuracy, either for TGCNN or other algorithms. The grasping position detection results of each algorithm using Gaussian-Mask and binary representations are listed in two columns as shown in Table~\ref{tab1} and~\ref{tab2}. As can be observed, by introducing Gaussian representation, the accuracy of all algorithms is greatly improved compared to binary representation. With binary representation, our algorithm does not always perform the best, because TGCNN is sensitive to the boundary, while other algorithms are not. It can be seen from Fig.~\ref{fig:effectcomparison} that TGCNN with binary representation locates the grasping center at the edge of the object, resulting in an unsatisfactory initial tactile detection position. However, with Gaussian representation, the grasping position is guided to the center, allowing tactile calibration to get a better initial position.

\textbf{(d) Multi-object Evaluation in Cluttered Scenes:} Besides predicting the optimal grasping of unseen objects, the robustness of TGCNN is also reflected in the ability to predict the grasping of multiple objects in cluttered scenes.  In the experiment, the training set contains 4,000 images with a single object, and another 1,000 images with multiple objects in the clutter are used for testing. In each test, we randomly change the object type, object position, camera position, and scene background in the scene (the scene backgrounds appeared in the training set). The comparison of different algorithms is shown in Fig.~\ref{fig:synthetic}(C). It can be seen that although TGCNN is only trained on a dataset with a single object, it can effectively predict the grasping position of multiple objects with better performance than other algorithms. 

\subsection{Exp. 2: Grasping Position Detection in Different Backgrounds}

\begin{figure*}
	\centering
	\includegraphics[width=0.98\textwidth]{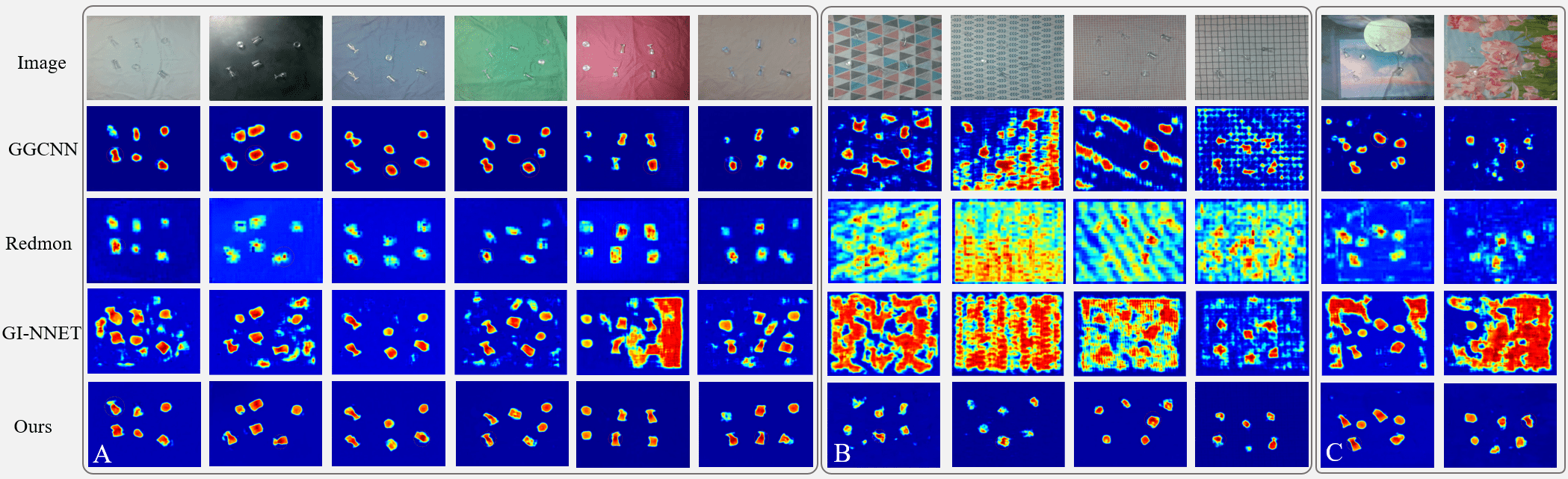}
	\caption{Comparison of the four algorithms for grasping position detection in different backgrounds: (A) Colored, (B) Patterned, and (C) Scenic backgrounds. } \label{fig:8.1}
\end{figure*}

To verify the grasping position detection performance of TGCNN in real scenes, we select 12 backgrounds with different features, including 6 colored backgrounds, 4 patterned backgrounds, and 2 scenic backgrounds, as shown in the first row in Fig.~\ref{fig:8.1}. The 6 objects in SimTrans12K are used for experiments. 4,000 synthetic data of two transparent objects are selected as the training set, 110 real data of 6 transparent objects as the test set which contains about 600 labels, and GOD is used to quantify the detection performance. The performance comparison of GGCNN, Redmon, GI-NNET, and TGCNN trained under the same dataset are shown in Fig.~\ref{fig:8.1} and Table~\ref{tab3}. The results reveal that all networks have good detection performance under a solid colored background (see Fig.~\ref{fig:8.1}(A)). 
While in the patterned and scenic backgrounds (see Fig.~\ref{fig:8.1}(B)), the GGCNN~\cite{morrison2018closing}, Redmon~\cite{redmon2015real}, and GI-NNet~\cite{shukla2022generative} algorithms produce more noise in the grasping position, whereas TGCNN still maintains good performance.

\begin{figure}[!t]
	\centering
	\includegraphics[width=0.48\textwidth]{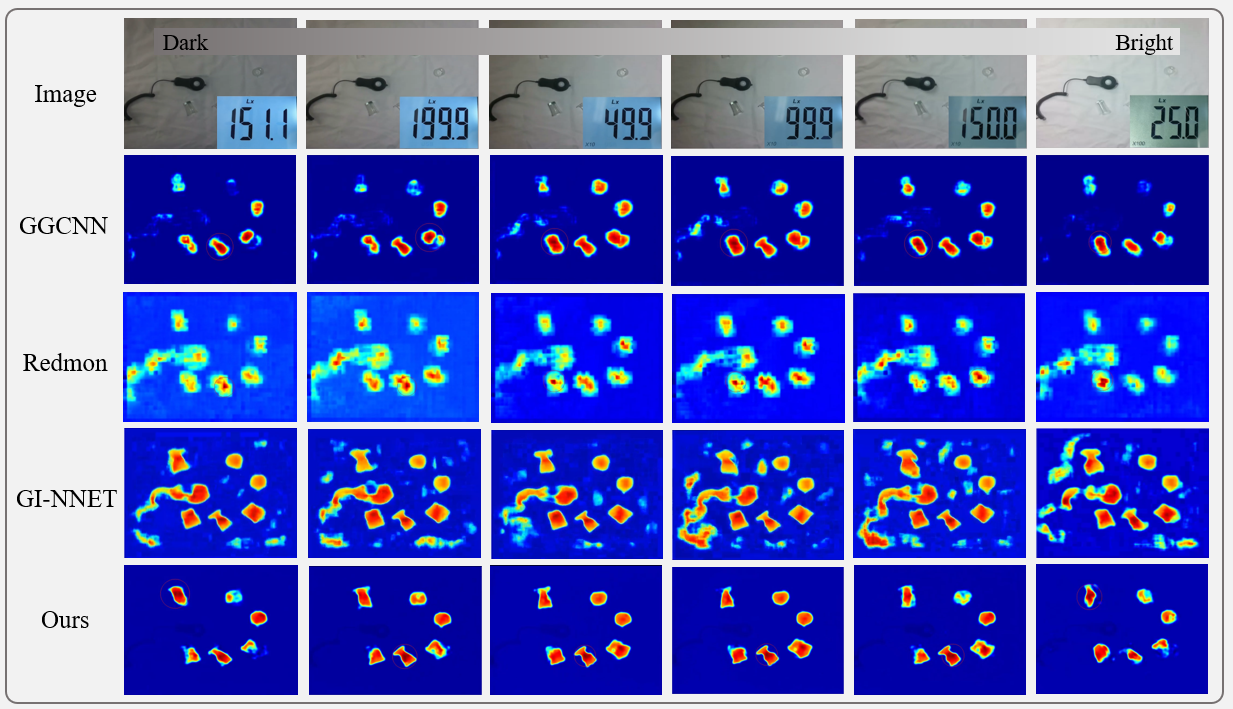}
	\caption{Comparison of the four algorithms for grasping position detection in different brightness.} \label{fig:light}
\end{figure}

\begin{table}[h]
	\centering	
	\caption{Detection results with real data}
	\begin{tabular}{m{1.8cm}<{\centering}|m{2.7cm}<{\centering}|m{2.6cm}<{\centering}}
		\hline
		Algorithm   & Accuracy (Different Backgrounds) &Accuracy (Different Brightness)   \\ \hline
		GGCNN~\cite{morrison2018closing}    & 51.8\% &  70.9\% \\    
		Redmon~\cite{redmon2015real}  &  73.2\%   &  73.2\% \\ 
             GI-NNET~\cite{shukla2022generative}  &      50.7\%     &   48.0\%   \\
		\textbf{Ours}      &     91.6\%   & 84.0\%    \\ \hline
	\end{tabular}
	\label{tab3}
\end{table}
Compared with GI-NNET, TGCNN has a larger number of parameters. Thanks to the application of residual layers \cite{he2016deep} and skip layer connections\cite{drozdzal2016importance}, we can increase the network depth while preventing the network from overfitting. In addition, TGCNN is a grasping network specially designed for jamming grippers, and we make some adjustments and optimizations in the number of layers and blocks of the network, so the TGCNN network has better detection results compared with GI-NNET.

\subsection{Exp. 3: Grasping Position Detection under Different Brightness}

Besides the background, the light condition is also an important factor affecting the detection accuracy. In this experiment, we test the impact of lightness on transparent object detection by changing the brightness (from 151 lux to  2,500 lux measured by a Lux Meter). 4,000 synthetic data of two transparent objects are selected as the training set and 50 real data of 6 transparent objects at different brightness as the test set. The detection results of the four networks are shown in Fig.~\ref{fig:light} and Table~\ref{tab3}. From the experimental results, we can see that the detection results of Redmon and TGCNN are relatively stable, but Redmon has more noise. The GGCNN and GI-NNET  networks have a stable detection effect when the brightness of the light is in the 199.9-999 lux interval. 
When the interval is exceeded the detection results will be affected, for example, some objects in the detection results of GGCNN will not be detected, and the detection results of GI-NNET will have more noise information. 

The experimental results show that the proposed method has good detection performance in real environments of different backgrounds and lighting conditions even though the network is trained using only synthetic data. In addition, we also test the influence of camera height and light position on the detection accuracy. It shows that when the camera height ranges between 35-120 cm, TGCNN maintains high detection accuracy. And under relatively uniform light conditions, the light position does not have an obvious impact on the detection performance.

\subsection{Exp. 4: Grasping and Classification on Planes with Complex Backgrounds}
To verify the effectiveness of the proposed transparent object grasping and classification framework, grasping and classification experiments are carried out. The selected objects are the same as in Fig.~\ref{fig:tactile_RGB}, including an angled wine glass and a smooth wine glass, a girdled water glass and a normal water glass, and a medicine bottle with a textured bottom and a smooth medicine bottle, which have slippery surfaces and similar shapes, and are difficult to both grasp and classify.
\begin{figure*}
	\centering
	\includegraphics[width=0.96\textwidth]{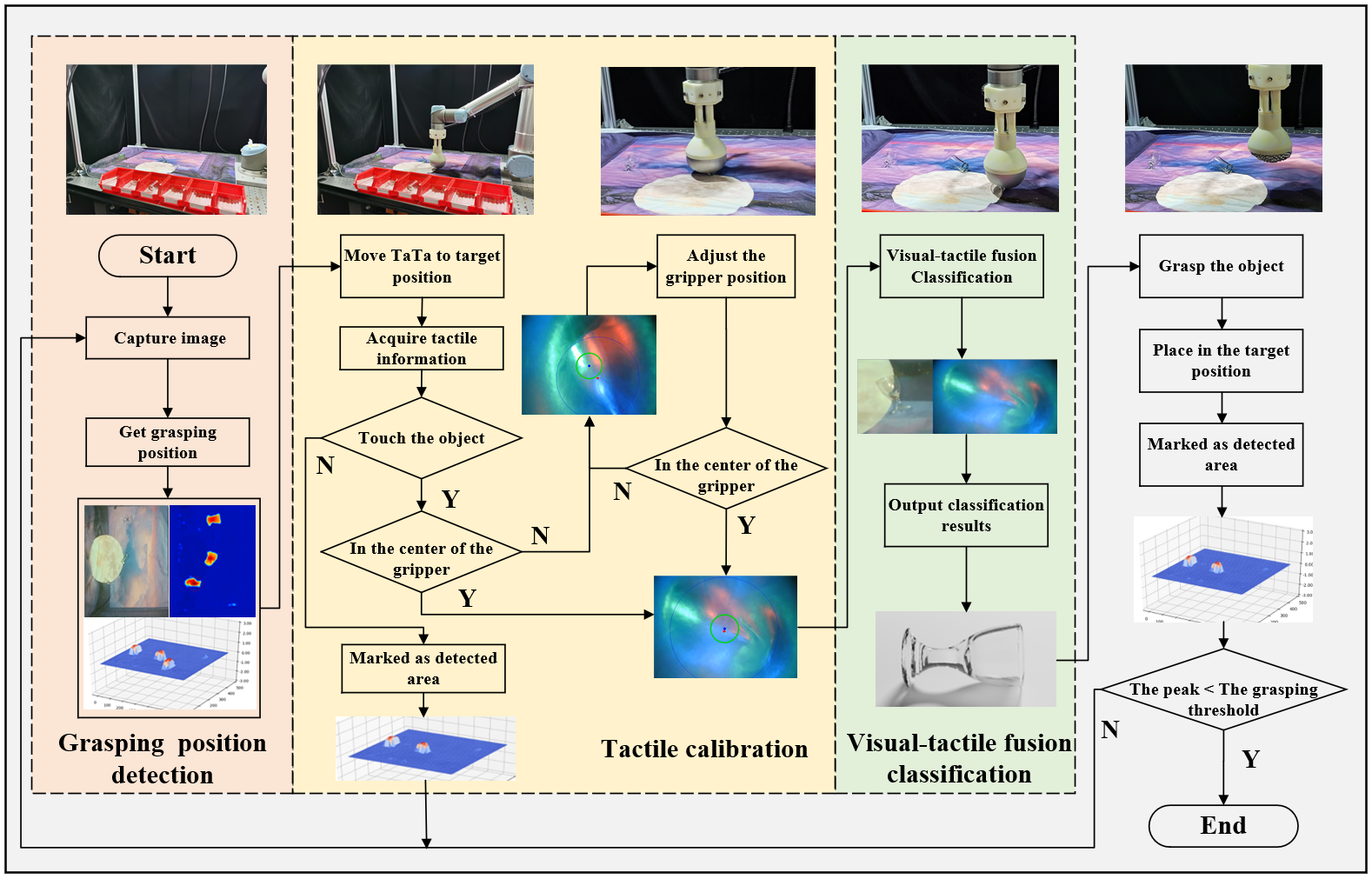}
	\caption{Flowchart of visual-tactile fusion based transparent object grasping and classification.} \label{fig:14}
\end{figure*}

The experimental procedure is shown in Fig.~\ref{fig:14}. Two backgrounds are used in the experiment---the pink background, which is relatively simple, and the moon background, which has various colors and complex textures. We compare the performance of GGCNN~\cite{morrison2018closing}, Redmon~\cite{redmon2015real}, GI-NNET~\cite{shukla2022generative}, and TGCNN. For each algorithm, we choose 3 objects randomly placed on the table each time and conduct 20 experiments on each background with a total of 60 grasping experiments on each background. 

\begin{table}[h]
	\centering	
	\caption{Experimental results of transparent object grasping and classification}
\begin{tabular}{m{1.7cm}<{\centering}|m{0.6cm}<{\centering}|m{0.6cm}<{\centering}|m{0.6cm}<{\centering}|m{0.6cm}<{\centering}|m{0.6cm}<{\centering}|m{0.6cm}<{\centering}}
    \hline
    Algorithm   &  \multicolumn{2}{m{1.8cm}<{\centering}} {Grasping Success Rate }  & \multicolumn{2}{|m{1.8cm}<{\centering}} {Classification Success Rate} & \multicolumn{2}{|m{1.8cm}<{\centering}} {No. of Tactile Calibration}    \\ \hline
    Background  & Pink & Moon & Pink & Moon& Pink & Moon\\ \hline

    GGCNN~\cite{morrison2018closing}    & 93\% &  90\%     & 95\% &  93\% & 2 &  4               \\    
    Redmon~\cite{redmon2015real}   & 93\% &  88\%     & 97\% &  95\% & 1 &  4    \\ 
    GI-NNET~\cite{shukla2022generative} & 87\% &  75\%     & 95\% &  93\% & 3 &  5    \\
   \textbf{Ours}        & 98\% &  93\%     & 98\% &  93\% & 1 &  2       \\ \hline
\end{tabular}
\label{tab4}
\end{table}

The experimental process is divided into three parts: grasping position detection, tactile calibration, and visual-touch fusion classification. In the grasping position detection stage, the image is acquired using RealSense D435i, and the transparent object grasping position and height are output using TGCNN. After getting to the grasping position, the gripping position will be adjusted using a tactile calibration algorithm. After reaching the optimal grasping position, the object will be classified using the visual-touch fusion classification algorithm and placed in the target location. Finally, we  compare the peak value of the output of the grasping position detection network with a preset threshold, repeat the above operation if it is greater than the threshold, and end the grasping if less than the threshold.

The experimental results are shown in Table~\ref{tab4}. The classification success rate in the table indicates the classification success rate in the case of successful grasping.
In addition, the number of tactile calibrations indicates the number of calibrations performed in each experiment in the case of successful grasping, which reflects the grasping position detection accuracy (the number of calibrations is less when the accuracy is higher).
The experimental results show that all four networks have good detection performance in the pure color background, while in complex backgrounds, TGCNN has a better performance in terms of grasping success rate and the tactile calibration number. Even in the case of poor grasping position detection, the tactile calibration algorithm in the framework still has a certain probability to achieve the grasping of transparent objects. In addition, we have also compared the detection effects of visual-tactile fusion classification and visual-only classification in the real experiments, and obtained results similar to the algorithm introduction section, with a detection accuracy improvement of 39\%.

\subsection{Exp. 5: Transparent Fragment Grasping}
Once a transparent object is broken, a large number of fragments will be produced, which have irregular shapes and various sizes and are difficult to grasp. To test the effectiveness of the visual-tactile fusion grasping framework, transparent fragment grasping experiments are performed, suggesting that tactile sensing has an important enhancement to the grasping success rate.

\begin{figure*}
	\centering
	\includegraphics[width=0.975\textwidth]{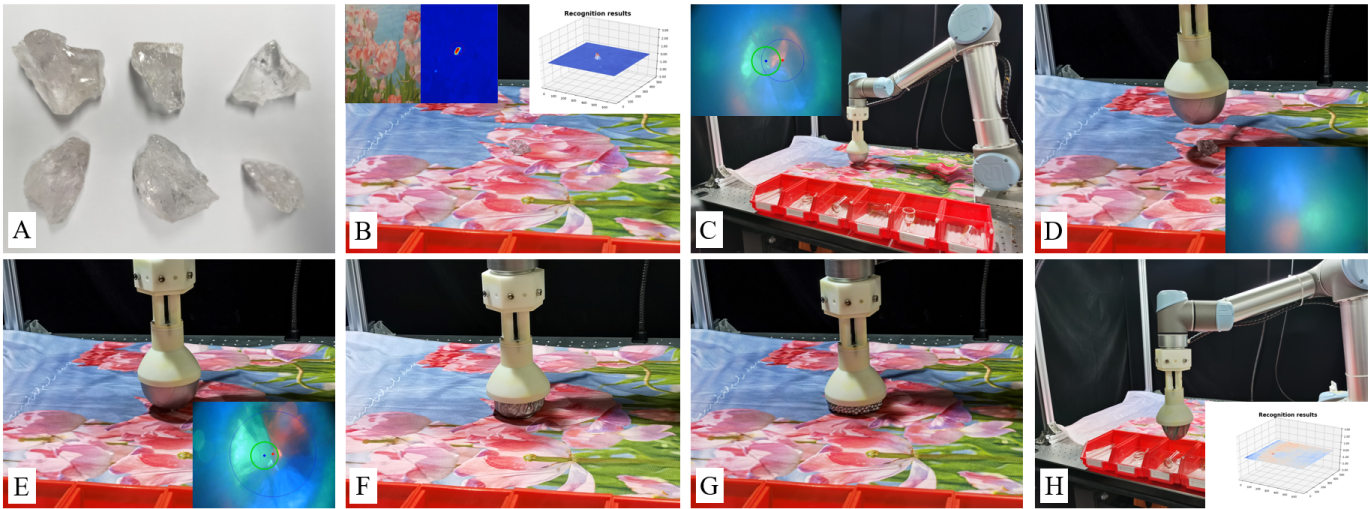}
	\caption{Transparent fragment grasping experiment based on visual-tactile fusion. (A) Transparent fragments. (B) Get the grasping position and height. (C) Contact with the object and detect the center of its contour are not in the gripper center. (D) Adjust the position of the gripper. (E) Touch the object again, whose contour center coincides with the gripper center. (F) (G) Grasp object. (H) Place the object in the specified position.} \label{fig:17}
\end{figure*}

The transparent fragments used in the experiment are shown in Fig.~\ref{fig:17}(A), which are some glass fragments with jagged surfaces, further increasing the difficulty of grasping position detection. The fragment grasping process omits the classification process compared to \textbf{Exp. 4}, but places higher demands on the grasping process, and the experimental process is shown in Fig.~\ref{fig:17}(B)-(H). To test the tactile calibration algorithm, grasping experiments with and without tactile calibration are performed. When tactile calibration is disabled, the gripper will grasp directly, without further adjustment of the grasping position. We compare the performance of GGCNN~\cite{morrison2018closing}, Redmon~\cite{redmon2015real}, GI-NNET~\cite{shukla2022generative}, and TGCNN. Each algorithm is tested in the yellow, grid, and flower backgrounds, and 20 visual-tactile fusion grasps and 20 direct grasps are performed in each background. 

It can be seen from Table~\ref{tab5} that in the yellow background, the detection accuracy remains high in direct grasping because the grasping position can be determined more accurately by a vision in the yellow background compared to grid and flower backgrounds. When TGCNN is adopted as the transparent object grasping position detection network, the tactile calibration method can improve the grasping success rate by 15\%, 50\%, and 45\% under the yellow, grid, and flower backgrounds separately and the overall grasping success rate by 36.7\%, showing the feasibility of the framework for transparent fragment grasping.

\begin{table}[h]
	\centering	
	\caption{Grasping success rates with and without tactile calibration}
\begin{tabular}{m{1.7cm}<{\centering}|m{0.65cm}<{\centering}|m{0.65cm}<{\centering}|m{0.65cm}<{\centering}|m{0.65cm}<{\centering}|m{0.65cm}<{\centering}|m{0.65cm}<{\centering}}
    \hline
    Algorithm   &  \multicolumn{3}{m{2.5cm}<{\centering}} {Tactile Calibration Grasping }  & \multicolumn{3}{|m{2.5cm}<{\centering}} {Direct Grasping}   \\ \hline
    Background  & Yellow &  Grid & Flower & Yellow &  Grid & Flower\\ \hline
    GGCNN \cite{morrison2018closing}    & 95\% &  85\%     & 80\% &     80\% & 30\% &  25\%               \\    
    Redmon    \cite{redmon2015real}  & 95\% &  75\%     & 80\% &     65\% & 15\% &  20\%   \\ 
    GI-NNET  \cite{shukla2022generative}  & 95\% &  85\%     & 65\% &      60\% &  30\% &  15\%   \\
 \textbf{Ours}     & 100\% &  90\%     & 95\% &    85\% & 40\% &  50\%       \\ \hline
\end{tabular}
\label{tab5}
\end{table}

\begin{figure*}
	\centering
	\includegraphics[width=0.97\textwidth]{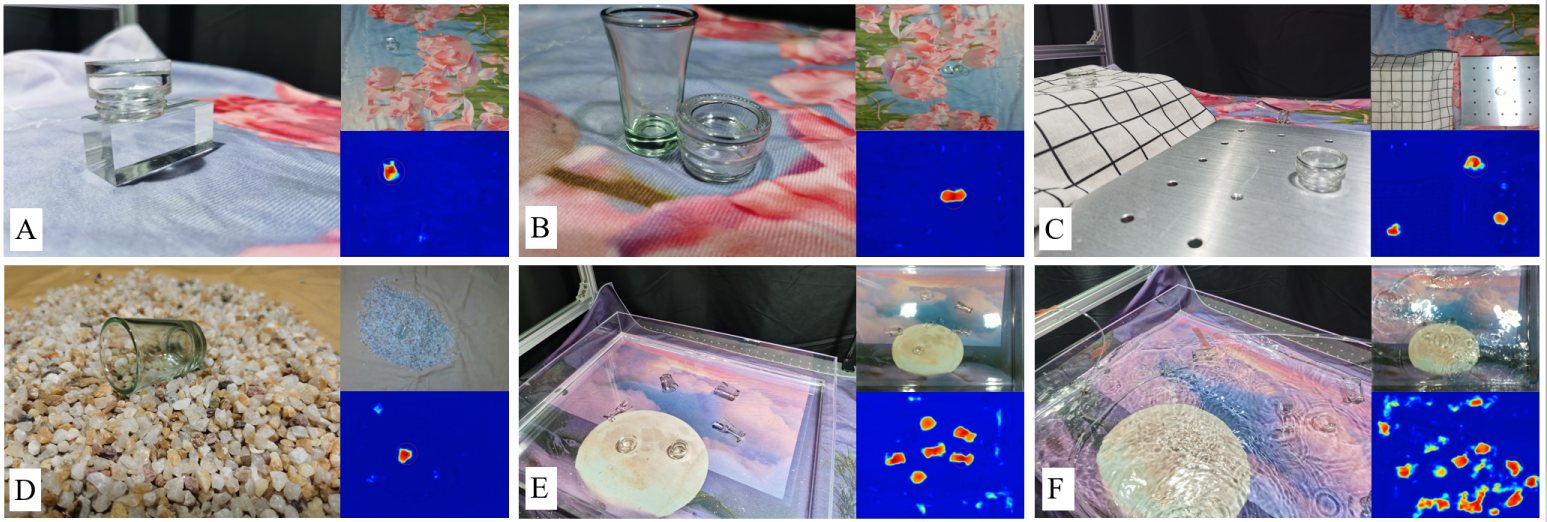}
	\caption{Transparent object in complicated scenes: (A) Overlapping, (B) Stacking, (C) Undulating, (D) Sand, (E) Underwater, and (F) Highly dynamic underwater scenes.} \label{fig:scenes}
\end{figure*}

\begin{figure*}
	\centering
	\includegraphics[width=0.96\textwidth]{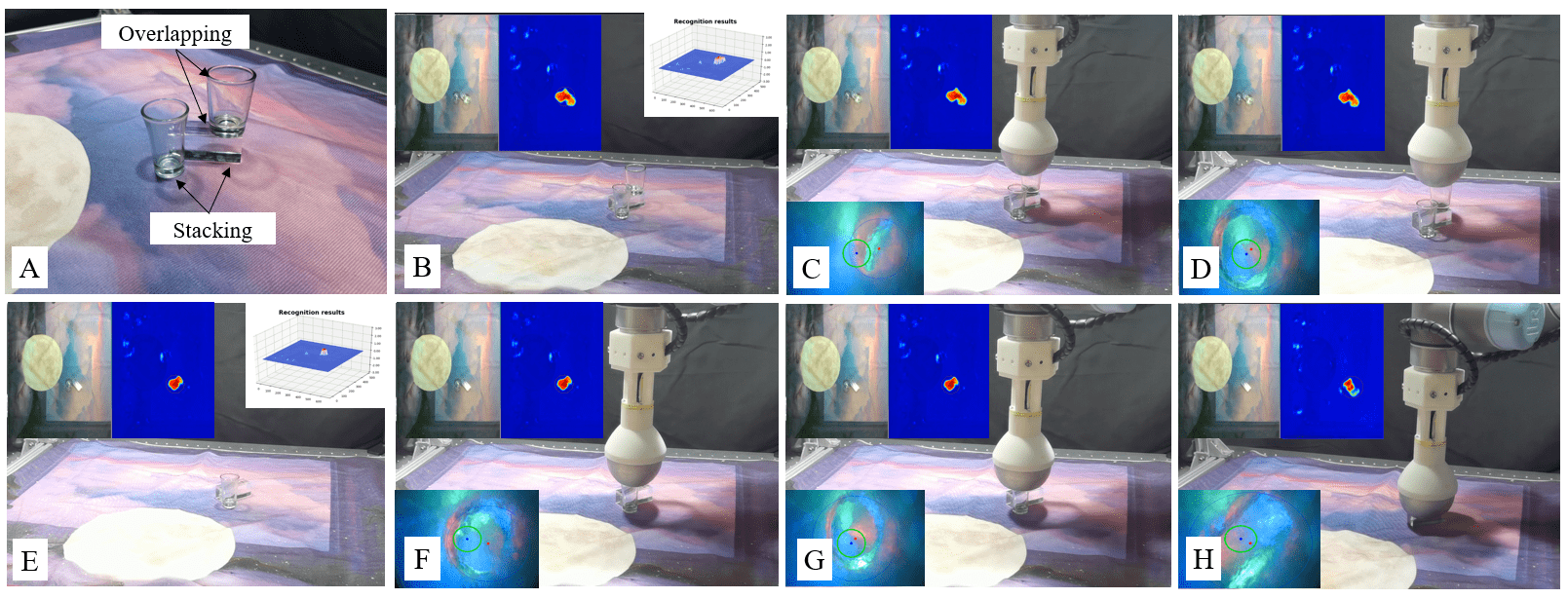}
	\caption{Transparent object grasping process in stacking and overlapping scenes. (A) Experimental setup. (B) Get the grasping position. (C) Use the THS module to search objects and contact the first object. (D) Adjust the grasping position with the tactile calibration module and grasp the object. (E) Get the new grasping position after completing the first object grasping. 
 (F)  Use the THS module to search objects and contact the second object. (G) Adjust the grasping position with the tactile calibration module and grasp the object. (H) Get the new grasping position and grasp the third object.} \label{fig:61}
\end{figure*}

\begin{figure*}
	\centering
	\includegraphics[width=0.96\textwidth]{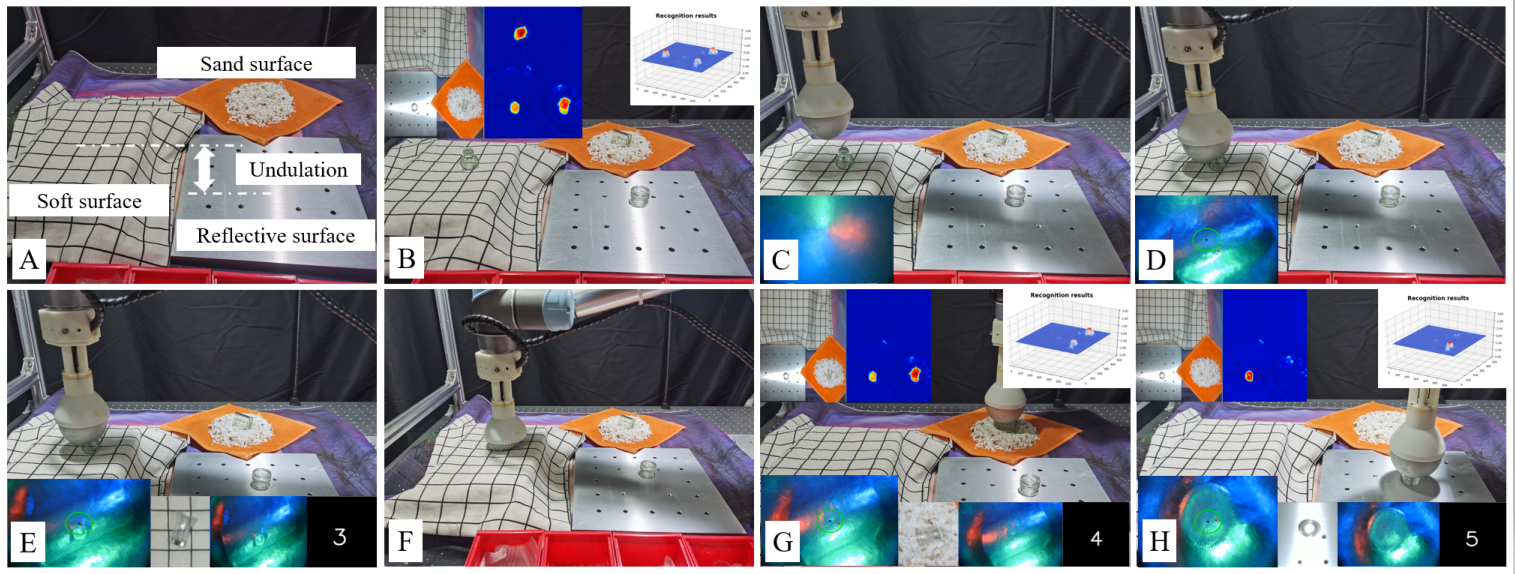}
	\caption{Transparent object grasping process in undulating and sand scenes. (A) Experimental setup. (B) Get the grasping position. (C) Arrive at the first object position. (D) Use the THS module to obtain the object height. (E) Adjust the grasping position with the tactile calibration module and use the visual-tactile fusion algorithm to classify the object. (F) Finish the first object grasp.  (G) Grasp and classify the second object.  (H) Grasp and classify the third object.} \label{fig:49}
\end{figure*}

\begin{figure*}
	\centering
	\includegraphics[width=0.96\textwidth]{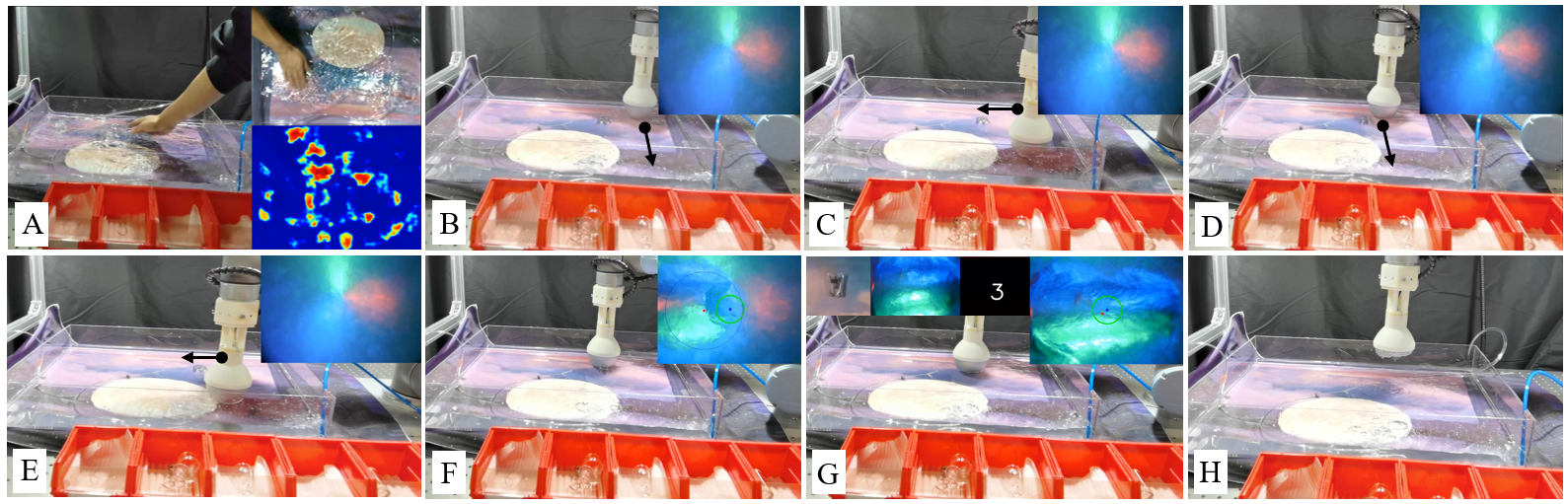}
	\caption{Transparent object grasping in highly dynamic underwater scenes. (A) Transparent object grasping position detection results. (B) (C) (D) (E) Explore the transparent object within a specific area using tactile perception. (F) Adjust the grasping position after contacting the object. (G) Visual-tactile fusion classification. (G) Finish grasping.} \label{fig:50}
\end{figure*}

\subsection{Exp. 6: Grasping in Irregular Scenes}

Compared to grasping on a plane, it is more challenging to grasp transparent objects in irregular scenes such as overlapping, stacking, undulating, and sand (see Fig.~\ref{fig:scenes}(A)-(D)), where the grasping position and height are difficult to obtain. To solve this problem, we add the THS module based on the previous framework to obtain the height where the object is located by tactile. As shown in Fig.~\ref{fig:scenes}(A)-(E), to verify the grasping effect of the THS module in irregular scenes, we conduct experiments on the grasping of transparent objects in the case of stacking and overlapping, as well as the grasping of transparent objects in special scenes such as undulating surfaces, sand, and  underwater.

To demonstrate the experimental process, we designed two representative scenes, the first with stacking and overlapping problems, the  second with undulating areas, reflective areas, and sand, as shown in Fig.~\ref{fig:61} and  Fig.~\ref{fig:49}. We conduct 20 grasping experiments in each scene, and the overall success rate can reach more than 90\%, which shows the feasibility of the method for grasping transparent objects on irregular planes. More experimental procedures can be found on the website 
\href{https://sites.google.com/view/visual-tactilefusion} {https://sites.google.com/view/visual-tactilefusion}.

\subsection{Exp. 7: Grasping in Visually Undetectable Scenes}

\begin{table}[h]
	\centering	
	\caption{ Experimental results for grasping in dynamic underwater}
	\begin{tabular}{m{3.5cm}<{\centering}|m{1.2cm}<{\centering}|m{1.2cm}<{\centering}|m{1.2cm}<{\centering}}
		\hline
		Exploration Step Length         & 5 cm & 10 cm  &  15 cm  \\ \hline
            Grasping Success Rate    & 90\%  & 65\%   &  25\%  \\ 
            Average Time Consumption  &  121s   & 78s  &  52s  \\ \hline
	\end{tabular}
	\label{tab6}
\end{table}

Finally, we test transparent object grasping in  highly dynamic underwater scenes, where the object becomes visually undetectable, as shown in Fig.~\ref{fig:scenes}(F). In this case, the touch-first grasp strategy is applied, which incorporates the TPE module. We assume that the object will not be moved by the water wave. The experimental procedure is shown in Fig.~\ref{fig:50}. Through the experiment, we find that the exploration step length (distance between two exploration  positions) of the gripper has a significant impact on the success rate of grasping, so we conduct a comparison experiment with exploration step lengths of 5 cm, 10 cm, and 15 cm, respectively, and 20 experiments are conducted for each step length. The results are shown in Table~\ref{tab6}, where the average time consumed refers to the time consumed to successfully find the object, and the failure cases are not counted. From the results, we can see that the smaller the step size, the higher the success rate of grasping, but also the more time consumed.

\begin{figure}[!t]
	\centering
	\includegraphics[width=0.48\textwidth]{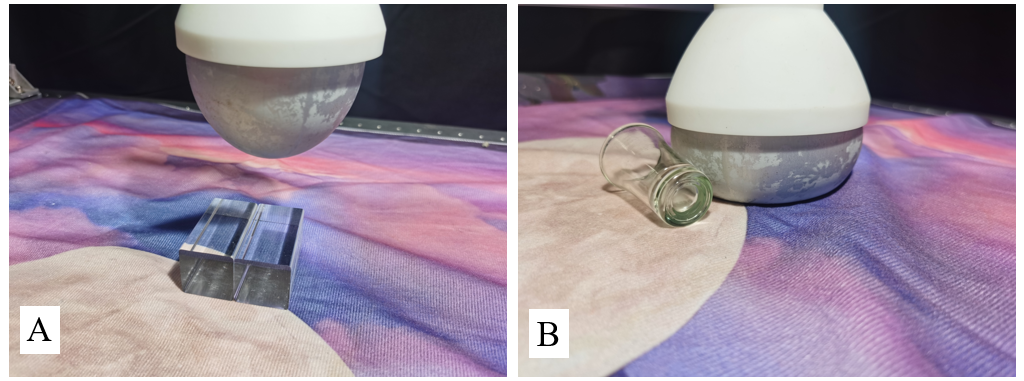}
	\caption{Some failed grasping trials. (A) Grasping trial for two flat transparent objects with the same height. (B) The transparent object collides with the gripper, thus causing the object to slide. } \label{fig:failure}
\end{figure}
Besides, we compare the grasping experiments in three environments, i.e., plane, irregular, and visually undetectable scenes. In each scene, we conduct 20 grasping experiments, and the average time consumption from the beginning to reach the appropriate grasping position is 22s, 32s, and 121s (the exploration step is 5 cm in a high-dynamic underwater scene) in three environments, respectively.

\subsection{Failed Trials and Limitations}

There are some failed trials in the experiments, as shown in Fig.~\ref{fig:failure}. The first case is to grasp two close transparent objects with the same height. It sometimes fails mainly because such transparent objects have not only the same texture but also similar height information. The second case is that if the grasping position detection error is close to the radius of sensing of the gripper, the edge of the gripper will easily collide with the transparent object and cause the object to slide. In this case, the grasping may also fail. However, as long as the deviation of the detected gripping position from the actual gripping position do not exceed the radius of the gripper, almost no slipping and failure will occur.

\section{Conclusion}
To solve the challenging problem of detecting, grasping, and classifying transparent objects, a visual-tactile fusion framework based on the synthetic dataset is proposed in this paper. First, we use the Blender simulation engine to render synthetic datasets rather than manually annotated datasets. 
Besides, we use Gaussian-Mask instead of the traditional binarized annotation to make the generation of the grasping position more accurate. To achieve grasping position detection for transparent objects, an algorithm named TGCNN is proposed and multiple comparative experiments are conducted, which show that the algorithm can achieve good detection under different backgrounds and lighting conditions even when trained with only synthetic datasets. 
Considering the grasping difficulty caused by the limitation of visual detection, we propose a tactile calibration method combined with the soft gripper TaTa to improve the grasping success rate by adjusting the grasping position with tactile information. The method improves the grasping success rate by 36.7\% compared to vision-only grasping.
Furthermore, to solve the classification problem of transparent objects in complex scenes, a transparent object classification method based on visual-tactile fusion is proposed, which improves the accuracy by 39.1\% compared to the vision-only based classification. In addition, to achieve transparent object grasping in irregular and visually undetectable scenes, we propose the THS and TPE modules, which can compensate for the problem of transparent object grasping in the absence of visual information. Extensive experiments are designed systematically and the results verify the effectiveness of the proposed framework in various complex scenarios, including stacking, overlapping, undulating, sand, underwater scenes, etc. We believe that the proposed framework can also be applied to object detection in low-visibility environments such as smoke and murky underwater, where tactile perception can compensate for the shortcomings of visual detection and improve classification accuracy by using visual-tactile fusion.
\bibliographystyle{ieeetr}
\bibliography{refe}

\end{document}